\documentclass[10pt,journal,compsoc]{IEEEtran}
\ifCLASSOPTIONcompsoc
  \usepackage[nocompress]{cite}
\else
  \usepackage{cite}
\fi
\ifCLASSINFOpdf
\else
\fi
\usepackage{amsmath,amsfonts}
\usepackage{amsthm,amssymb}
\usepackage{mathrsfs}
\usepackage{algorithmic}
\usepackage{algorithm}
\usepackage{array}
\usepackage{hyperref}

\ifCLASSOPTIONcompsoc
\usepackage[caption=false, font=normalsize, labelfont=sf, textfont=sf]{subfig}
\else
\usepackage[caption=false, font=footnotesize]{subfig}
\fi

\usepackage{textcomp}
\usepackage{stfloats}
\usepackage{url}
\usepackage{verbatim}
\usepackage{graphicx}

\usepackage{cite}

\usepackage{soul}
\usepackage{bm}
\usepackage{booktabs}
\usepackage{txfonts}
\usepackage{booktabs}
\usepackage{multirow}
\usepackage{caption3}
\usepackage{caption}
\usepackage{xcolor}
\usepackage{float}

\usepackage{pifont}
\usepackage{threeparttable}

\begin{document}


%
\title{MetaEarth: A Generative Foundation Model for Global-scale Remote Sensing Image Generation}
%
%
%
%

\author{Zhiping Yu, Chenyang Liu, Liqin Liu, Zhenwei Shi,~\IEEEmembership{Senior Member,~IEEE}, \\
and Zhengxia Zou$^*$,~\IEEEmembership{Senior Member,~IEEE}
\IEEEcompsocitemizethanks{
\IEEEcompsocthanksitem Zhiping Yu, Chenyang Liu, Liqin Liu, and Zhenwei Shi are with the Image Processing Center, School of Astronautics, Beihang University, Beijing 100191, China, and with the State Key Laboratory of Virtual Reality Technology and Systems, Beihang University, Beijing 100191, China, and also with the Shanghai Artificial Intelligence Laboratory, Shanghai 200232, China.
\IEEEcompsocthanksitem Chenyang Liu is also with Shen Yuan Honors College, Beihang University, Beijing 100191, China.
\IEEEcompsocthanksitem Zhengxia Zou is with the Department of Guidance, Navigation and Control, School of Astronautics, Beihang University, Beijing 100191, China. 
\IEEEcompsocthanksitem Corresponding author: Zhengxia Zou (zhengxiazou@buaa.edu.cn)

}
\thanks{The work was supported by the National Natural Science Foundation of China under Grant 62125102 and 62471014, the National Key Research and Development Program of China (Grant No. 2022ZD0160401), the Beijing Natural Science Foundation under Grant JL23005, and the Fundamental Research Funds for the Central Universities.}}

%
%

\markboth{Journal of \LaTeX\ Class Files,~Vol.~XX, No.~XX, XXX~XXXX}%
{Shell \MakeLowercase{\textit{et al.}}: Bare Demo of IEEEtran.cls for Computer Society Journals}
%



\IEEEtitleabstractindextext{%
\begin{abstract}
The recent advancement of generative foundational models has ushered in a new era of image generation in the realm of natural images, revolutionizing art design, entertainment, environment simulation, and beyond. Despite producing high-quality samples, existing methods are constrained to generating images of scenes at a limited scale. In this paper, we present MetaEarth - a generative foundation model that breaks the barrier by scaling image generation to a global level, exploring the creation of worldwide, multi-resolution, unbounded, and virtually limitless remote sensing images. In MetaEarth, we propose a resolution-guided self-cascading generative framework, which enables the generating of images at any region with a wide range of geographical resolutions. To achieve unbounded and arbitrary-sized image generation, we design a novel noise sampling strategy for denoising diffusion models by analyzing the generation conditions and initial noise. To train MetaEarth, we construct a large dataset comprising multi-resolution optical remote sensing images with geographical information. Experiments have demonstrated the powerful capabilities of our method in generating global-scale images. Additionally, the MetaEarth serves as a data engine that can provide high-quality and rich training data for downstream tasks. Our model opens up new possibilities for constructing generative world models by simulating Earth’s visuals from an innovative overhead perspective.
\end{abstract}

}

\maketitle

\IEEEdisplaynontitleabstractindextext

%
\IEEEpeerreviewmaketitle

\IEEEraisesectionheading{\section{Introduction}\label{sec:introduction}}

%
%
%
%

 

\IEEEPARstart{T}{he} ongoing advancement of generative foundation models has propelled natural image generation techniques to a new level\cite{10230895}. These encompass text-to-image generation~\cite{txt2img1, txt2img2, stable-diffusion},  video generation~\cite{video1, video2}, image translation~\cite{img2img1, img2img2, img2img3}, image editing~\cite{imgediting1, imgediting2}, etc, contributing to many practical applications, including art design~\cite{artdesign1}, training image augmentation~\cite{dataaugmentation, azizi2023synthetic, wu2023datasetdm}, and building generative world models~\cite{zhu2024sora, hu2023gaia, genie}.

\begin{figure*}
	\centering
	\includegraphics[width=1.0\linewidth]{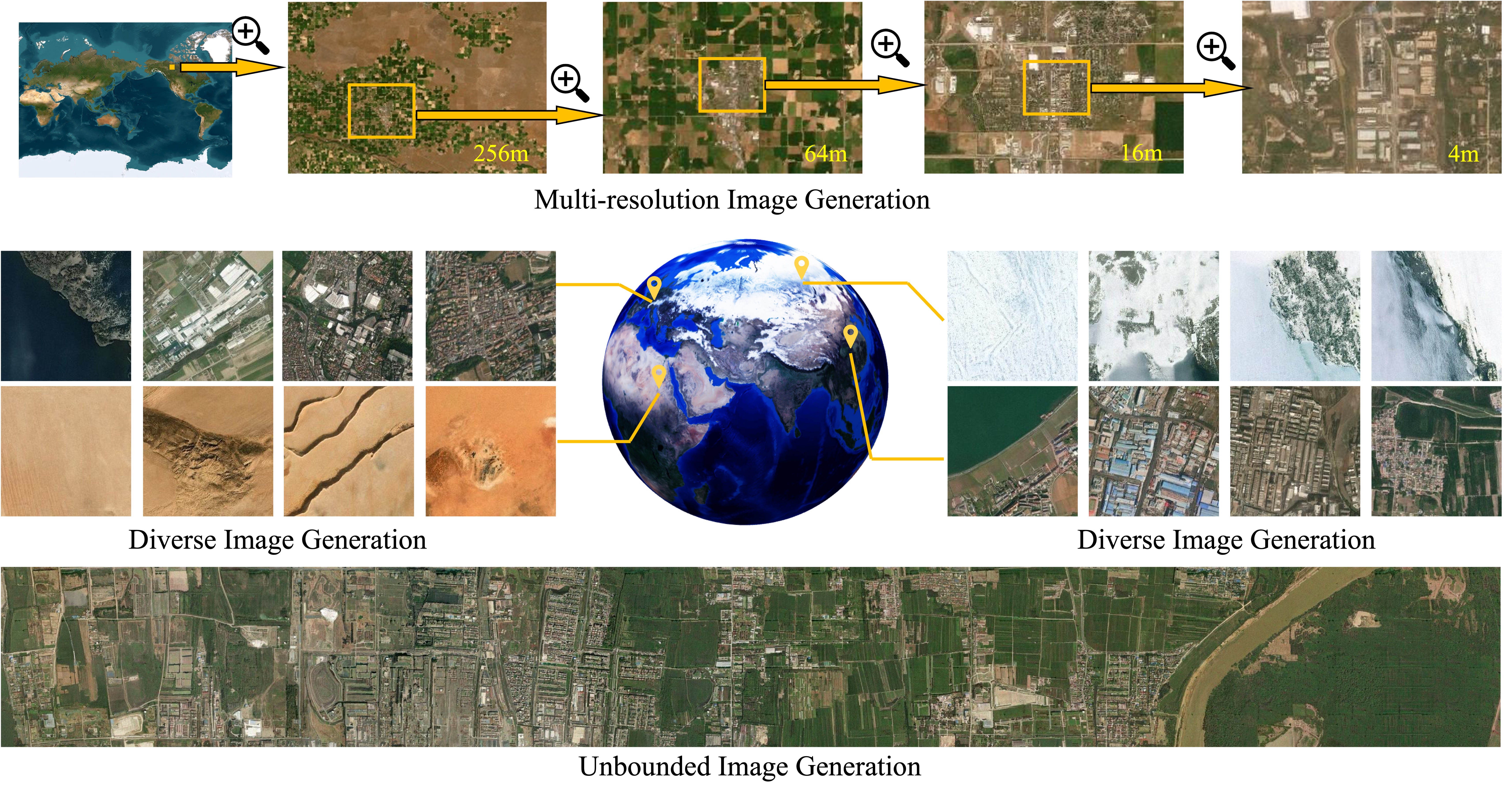}
	\caption{
    We propose MetaEarth, a generative foundation model that simulates Earth’s visuals from an overhead perspective. MetaEarth shows powerful capabilities on generating worldwide, multi-resolution, unbounded, and virtually limitless remote sensing images. For more visualization and animated results, please refer to our project page: \url{https://jiupinjia.github.io/metaearth/}.
 }
	\label{fig:teaser}
\end{figure*}

Despite the recent progress in image generation quality and diversity, the scale of the scene generated is still constrained to daily human activity scenarios, resulting in a single generated image with limited resolution and limited information capacity. In this paper, we aim to push the boundaries of generative models by extending image generation capabilities from localized human daily scenes to a global scale. We employ an overhead remote sensing perspective as a novel viewpoint to study this problem and present a foundational generative model MetaEarth. Our methods innovatively propose a self-cascading framework and a noise sampling strategy, enabling MetaEarth to generate worldwide, multi-resolution, unbounded, and virtually limitless remote sensing images. Fig.~\ref{fig:teaser} showcases some results of our model.

In our research, we primarily face three challenges. The first challenge is about model capacity. Generating world-scale images involves a wide range of geographical features, including cities, forests, deserts, oceans, glaciers, and snowfields. Even only within the category of cities, remote sensing images vary significantly across different latitudes, climates, and cultural landscapes. This diversity imposes substantial demands on the capacity of the generative model. The second challenge involves the generation of images with controllable resolution. In the overhead imaging process, different imaging altitudes correspond to images with different resolutions, resulting in significant differences in the details of geographical features. However, there are currently few studies on image generation with controllable resolution. Therefore, generating remote sensing images at specified resolutions (meters per pixel) from any geographic location is a highly challenging task. The third challenge involves the generation of unbounded images. Unlike everyday natural images, remote sensing images are uniquely characterized by their vast size, often measuring tens of thousands of pixels in length and width. Current natural image generation algorithms are generally limited to producing images of sizes such as 512$\times$512 pixels or 1024$\times$1024 pixels. Therefore, in addition to addressing the first two challenges, generating continuous, unbounded images of arbitrary size remains a significant unsolved problem in existing methods.

To address the challenge of model capacity, we constructed a world-scale generative foundation model with over 600 million parameters based on the denoising diffusion paradigm. To train the model across diverse spatial resolutions globally, we collected a large-scale remote sensing image dataset comprising images and their geographical information (latitude, longitude, and resolution) from multiple spatial resolutions that cover most regions globally. 
To generate cross-resolution and controllable remote sensing images, we propose a novel self-cascading generative framework. Different from previous single-stage generation methods~\cite{stable-diffusion, DALLE2}, the proposed framework generates multi-resolution images from a given geographical location in a recursive manner. In different cascading stages, images are generated sequentially from low to high resolutions. Specifically, we constructed a unified model across different cascading stages, where in each stage, the previously generated low-resolution image, as well as the geographical resolution, are embedded as controllable generative variables to guide the generation of images at higher resolutions. As the cascading stages accumulate, the generated images manifest diversity in both resolution and content, enabling the generation of varied parallel scenarios. 

To generate unbounded and arbitrary-sized images, a straightforward idea is to stitch multiple generated image blocks into a larger image. However, due to the inherent stochasticity of generative models, directly stitching independently generated image blocks can result in visually discontinuous seams between the blocks. To solve this problem, we design a novel noise-sampling strategy for denoising diffusion models by analyzing the generation conditions and initial noise. The proposed strategy ensures consistency in style and semantics between generated image tiles, enabling the generation of arbitrary-sized images from smaller pieces. 

We conducted comprehensive experimental analyses on MetaEarth, including both quantitative and qualitative evaluations. Experiment results have demonstrated that MetaEarth achieves the generation of high-quality images with broad coverage, multiple resolutions, and diverse image contents. The ablation studies also suggest the effectiveness of the above-mentioned methodology design. Additionally, MetaEarth can serve as a data engine, providing high-quality and diverse training data for downstream tasks. In our experiments, choosing image classification as an example, we demonstrated that the high-quality generated samples provided by MetaEarth can significantly improve classification accuracy. 

The contributions of this paper are summarized as follows:

\begin{itemize}
\item  
We propose MetaEarth, a novel generative foundation model for remote sensing, exploring the creation of worldwide, multi-resolution, unbounded, and virtually limitless remote sensing images. To the best of our knowledge, MetaEarth is the first foundational model to extend image generation to a global scale.

\item 
We proposed a resolution-guided self-cascading generative framework and a noise sampling strategy to address the challenges faced by previous foundational models in generating cross-resolution and continuous unbounded images.

\item 
As a generative data engine, MetaEarth has the potential to provide virtual environments and realistic training data support for various downstream tasks in remote sensing and Earth observation. Our model also opens up new possibilities for constructing generative world models~\cite{zhu2024sora, hu2023gaia, genie} by simulating Earth’s visuals from an innovative overhead (bird’s-eye view) perspective.
\end{itemize}

\section{Related work}

In this section, we briefly review the recent advancements in visual generative models, probabilistic diffusion models for image generation, and image generation models in the field of remote sensing.

\subsection{Visual Generative Models}

In recent years, image generation technology has attracted considerable attention in computer vision and machine learning. Image generation models aim to capture underlying patterns from real image data distributions and synthesize realistic images. With the development of deep learning-based generative models, significant progress has been made in recent years. Existing image generation models fall into four main categories: Generative Adversarial Networks (GANs)~\cite{gan,karras2019style,esser2021taming}, Variational Autoencoders (VAEs)~\cite{vae, van2017neural}, Flow-based models~\cite{flow}, and Diffusion models~\cite{ddpm, improvedddpm, stable-diffusion}.

GANs, consisting of a generator and a discriminator trained simultaneously, produce high-quality synthetic data resembling real data but struggle with diversity and stable training due to mode collapse. VAEs, inspired by variational Bayes inference, map input data to latent space, generating diverse samples and providing uncertainty estimates, though often producing blurry reconstructions. Flow-based Models, which learn invertible transformations for exact likelihood computation and efficient sampling, face challenges in modeling complex data distributions, making them unsuitable for global image generation. Diffusion Models, using a noise-corruption and reverse process, have shown remarkable performance in generating high-quality images despite slow generation speeds. In this paper, we leverage diffusion models to generate global remote sensing images.

\subsection{Diffusion-Based Image Generation}

Diffusion-based image generation can be divided into two categories~\cite{diffusionsurvey}: unconditional generation and conditional generation.

An unconditional generation model refers to a model that generates images without depending on any form of control, operating entirely in a random sampling manner. The most representative work is the Denoising Diffusion Probability Model (DDPM) proposed by Ho et al.~\cite{ddpm}. The DDPM builds upon the theoretical foundation established by Sohl-Dickstein et al.~\cite{sohl2015deep}, operating by reversing a diffusion process. This method disrupts real data during the forward process, adding noise to the data in small steps until it becomes Gaussian noise, then, learns to reverse these steps to generate data from noises. Based on DDPM, Nichol et al.~\cite{improvedddpm} propose learning variance to improve model performance. This design not only enhances the quality of generated images but also accelerates sampling speed. To further accelerate sampling, Song et al. propose the Denoising Diffusion Implicit Model (DDIM)~\cite{ddim}, modeling the diffusion process as a non-Markovian process. Without requiring adjustments to the training process of DDPM, the DDIM generation framework significantly speeds up the sampling procedure with only a small impact on quality. Additionally, the DDIM framework enables a deterministic generation process, ensuring that generated images depend solely on the initial noise of the reverse process.

The conditional diffusion model refers to a class of models that generate images based on additional signals, building upon the unconditional diffusion model. According to the type of input conditions, they can be divided into class-to-image generation~\cite{cascaded}, image-to-image synthesis~\cite{img2img1, img2img2, sr3,zhang2023adding}, text-to-image generation~\cite{stable-diffusion, imagen}, image editing~\cite{imgediting1, imgediting2} etc. Since DDPM's reverse process directly operates in pixel space, more computation and memory are required to predict high-dimension noise, making it challenging for the model to generate high-resolution images. There are two main approaches to addressing this issue. The first is cascaded generation~\cite{cascaded, imagen}, where low-resolution images are generated first and then used as conditional inputs to generate higher-resolution images sequentially. The second approach is latent space generation~\cite{stable-diffusion}, using an encoder-decoder network structure to compress the image into a lower-resolution latent space, followed by a denoising diffusion process within the latent space.

In this paper, we adopt a cascaded generation strategy, but with a novel approach that differs from previous methods. Firstly, previous models require designing and training multiple models separately for each resolution stage, which inevitably increases training costs. In contrast, our model shares the same network weights across different stages, using resolution embedding encoding to guide the learning of variable-resolution information. Additionally, while previous methods typically impose size constraints on input and output images at each stage, our approach does not have such restrictions, offering greater flexibility.

\subsection{Remote Sensing Image Generation}

In the field of remote sensing, image generation models have recently garnered increasing attention. As shown in Table \ref{tab:rs-gen}, previous research on this topic mainly focuses on text-to-image and image-to-image tasks.

\begin{table}[!ht] 
\renewcommand{\arraystretch}{1.3}
\captionsetup{justification=centering, labelsep=newline, font=small}
\caption{Comparison of existing generative models for remote sensing images.}
\label{tab:rs-gen}
\centering
\begin{tabular}{m{60pt}<{\centering}m{50pt}<{\centering}m{20pt}<{\centering}m{30pt}<{\centering}m{30pt}<{\centering}}
	\toprule
	Method & Control Signal & Global-scale & Resolution Range & Image Size (pixels) \\
	\midrule
        DiffusionSat~\cite{txt2rs2} & text, metadata & {$\times$} & limited & 512$\times$512 \\
        Crs-diff~\cite{txt2rs3} & text, metadata image & {$\times$} & limited & 224$\times$224 \\
        RSFSG-Diff~\cite{sem2rs1} & semantic map & {$\times$} & fixed & 256$\times$256 \\
        MapSat~\cite{map2rs} & map & {$\times$} & fixed & 256$\times$256 \\
        MetaEarth (ours) & image, metadata & {$\checkmark$} & multi-resolution & unbounded \\
	\bottomrule
\end{tabular}
\end{table}

For the text-to-image generation task, one common approach is to fine-tune latent diffusion models~\cite{stable-diffusion} directly based on remote sensing text-image datasets~\cite{txt2rs1, txt2rs2, txt2rs3}. The image-text data pairs are usually obtained by a pre-trained image captioning model and then are refined manually assisted by GPT-4. In addition to the direct fintuning, Sebaq et al.~\cite{txt2rs4} propose a generation framework containing two individual diffusion models, where the first model generates images from prompts and the second one upscales the generated images.

The image-to-image generation task is primarily divided into two categories: layout-to-image~\cite{sem2rs2} and modality transfer~\cite{10327767, chen2024spectral}. The former's input conditional images typically include maps~\cite{map2rs} and semantic layouts~\cite{sem2rs1, sem2rs3}, while the latter aims to achieve the transformation between multimodal data such as RGB images, infrared images, SAR images, and hyperspectral images.

Despite some attempts at image generation in the field of remote sensing, previous methods have been typically trained on small-scale or limited datasets, lacking diversity in land cover types and resolutions. Additionally, these methods have only been capable of generating remote sensing images up to 512$\times$512 pixels in size, with few attempts made at generating images for larger-scale scenes. In this paper, we thoroughly explore these issues by designing and training a foundational generative model for remote sensing, achieving promising results.

\begin{figure*}
	\centering
	\includegraphics[width=1.0\linewidth]{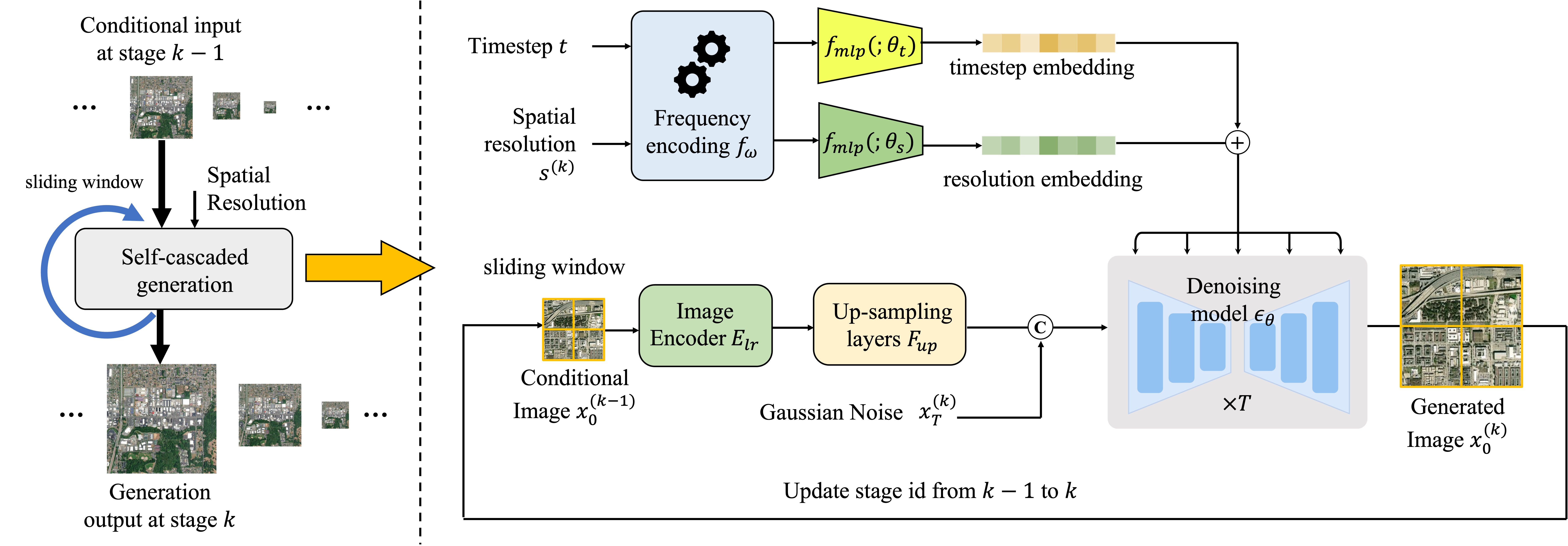}
	\caption{
    The overall structure of MetaEarth. We propose a resolution-guided, self-cascading framework which is capable of generating scenes and resolutions for any global region. The generation process unfolds in multiple stages, starting with low-resolution images and advancing to high-resolution images. In each stage, the generation is conditioned on the low-resolution images generated in the preceding stage and their spatial resolution.
 }
	\label{fig:method_overview}
\end{figure*}

\subsection{Remote Sensing Foundation Model}
The foundation models have recently been preliminarily studied in the field of remote sensing, emerging as a new research direction. These models are per-trained on large amounts of data and can be adapted to various downstream tasks through fine-tuning, few-shot learning, and zero-shot learning. Current research on foundation models in remote sensing mainly focuses on two aspects: vision-centric foundation models and vision-language foundation models.

In earlier years, vision-centric foundation models in remote sensing are typically trained using supervised learning. However, these models are constrained by the limited availability of supervised training data. Recently, self-supervised techniques have become mainstream, as they can leverage large amounts of unlabeled data. The self-supervised pre-trained foundation models are designed to learn robust feature representations from unlabeled remote sensing images. They can then be transferred to various downstream tasks, such as image classification, object detection, and segmentation, resulting in improved performance. These models predominantly rely on the Masked Autoencoders (MAE) architecture\cite{MAE}~\cite{sun2022ringmo,hong2024spectralgpt,bastani2023satlaspretrain} and are tailored to exploit the distinctive features inherent in remote sensing imagery, including image content~\cite{9956816}, multi-scale properties~\cite{Reed_2023_ICCV}, temporal dynamics~\cite{cong2022satmae}, and multispectral attributes~\cite{noman2024rethinking, hong2024spectralgpt}. Besides, the improvement of the model's structure and training paradigm~\cite{mendieta2023towards} is used to enhance the model's capacity for the representation of remote sensing data.

Remote sensing vision-language foundation models are trained on a large amount of image-text paired data. Some methods~\cite{liu2024remoteclip,zhang2023rs5m,mall2023remote} adopt contrastive learning, which aims to train an image encoder and a text encoder to align the visual and linguistic domains. The learned aligned vision-language representations can be used for many downstream tasks, such as zero-shot classification~\cite{liu2024remoteclip}, image-text retrieval~\cite{yuan2023parameter,silva2024large}, image captioning~\cite{Liu_2022,liu2024rscama}, and change captioning~\cite{RSICCformer,liu2023decoupling}. For example, Zhang \textit{et al.}~\cite{zhang2023rs5m} filtered remote sensing image-text pairs from publicly available large datasets to build the RS5M dataset and fine-tuned the CLIP~\cite{radford2021learning} model to obtain a GeoRSCLIP model. Recently, autoregressive pre-training has been widely used for training vision-language models, such as~\cite{kuckreja2023geochat,zhan2024skyeyegpt,zhang2024earthgpt,wang2024skyscript}. Most methods utilize CLIP to extract visual features and design connectors to map these visual features into large language models for predicting corresponding text. They typically start with image-text pre-training and then further refine the model through instruction fine-tuning.

Although the aforementioned work has made preliminary explorations in foundational models for remote sensing, research on generative foundational models in this field is still relatively scarce. In this paper, we propose MetaEarth, a novel generative foundation model for global-scale remote sensing image generation, which expands the research boundaries of current remote sensing foundation models.

\section{METHODOLOGY}

In this section, we will first revisit the Denoising Diffusion Probabilistic Model. Then, we introduce the proposed self-cascading generation framework and unbounded image generation method.

\subsection{Revisit of Denoising Diffusion Probabilistic Model}

The Denoising Diffusion Probabilistic Model (DDPM)~\cite{ddpm} is a type of generative model that leverages a diffusion process for image generation. It involves a \textit{forward process}, where the image data is gradually corrupted with noise, and a \textit{reverse process}, where a neural network learns to denoise the noisy image step-by-step to recover its original content. Through the \textit{reverse process}, high-quality images can be generated by iteratively refining noisy samples, leading to impressive results in various image synthesis tasks.

Suppose $x_0$ denotes a clean image. During the \textit{forward process}, a set of Gaussian noises are added to $x_0$ gradually according to a variance schedule ${\beta_1, \beta_2, ..., \beta_T}$ over $T$ time-steps:
\begin{subequations}
\label{Equation: forward process}
	\begin{align}
		q({x_t}|{x_{t-1}})&:=\mathcal{N}({x_t};\sqrt{1-{\beta_t}}{x_{t-1}},{\beta_t}I)\\
  		q({x_1,x_2,...,x_T}|{x_0})&:=\prod_{t=1}^{T}{q({x_t}|{x_{t-1}})}
	\end{align}
\end{subequations}
where $q({x_1,x_2,...,x_T}|{x_0})$ represents the joint probability distribution of $x_1,x_2,...,x_T$ conditioned by $x_0$. $q({x_t}|{x_{t-1}})$ represents the conditional distribution from $x_{t-1}$ to $x_{t}$ in a one-step diffusion processing. Since the \textit{forward process} of DDPM operates as a Markov chain, according to Eq.~(\ref{Equation: forward process}), by setting ${{\alpha_t}=\prod_{i=1}^{t}(1-\beta_i)}$, the conditional distribution of ${x_t}$ can be directly represented as a multivariate Gaussian distribution:
\begin{equation}
\label{Equation: qsample}
q({x_t}|{x_0})=\mathcal{N}(x_t;\sqrt{\alpha_t}{x_0},(1-{\alpha_t})I)     
\end{equation}

By utilizing the reparameterization trick and introducing a noisy image layer $\epsilon$ sampled from a Gaussian distribution, ${{\epsilon}\sim{\mathcal{N}(\boldsymbol{0},I)}}$, the relationship between ${x_t}$ and ${x_0}$ can be written as a one-step blending operation:
\begin{equation}
\label{Equation: xt}
x_t=\sqrt{\alpha_t}{x_0}+\sqrt{1-\alpha_t}\epsilon 
\end{equation}
Given the \textit{forward process} described by Eq.~(\ref{Equation: xt}), one can easily prove that if the scheduling parameters ${\beta_t}$ are properly set, when $T$ is sufficiently large, the distribution of ${x_T}$ can be approximated to a standard multivariate Gaussian distribution, i.e., ${{q(x_T|x_0)}\sim\mathcal{N}(0,I)}$.

In the \textit{reverse process}, starting with ${p(x_T)\sim{\mathcal{N}({x_T};\boldsymbol{0},I)}}$, we aim at recovering the $x_0$ step-by-step through a denoising processing. In each denoising step, a trainable network with parameters ${\theta}$ is trained to estimate the distribution ${q(x_{t-1}|x_t)}$:
\begin{subequations}
\label{Equation: reverse process}
	\begin{align}
	p_\theta({x_{t-1}}|{x_t})&:=\mathcal{N}({x_{t-1}};{\mu_\theta(x_t,t)},{\Sigma_\theta(x_t,t)})\\
 p_\theta({x_0,x_1,x_2,...,x_T})&=p(x_T)\prod_{t=1}^{T}{p_\theta({x_{t-1}}|{x_t})}
	\end{align}
\end{subequations}
According to~\cite{ddpm}, the model can be trained by optimizing the Variational Lower Bound(VLB):
\begin{subequations}
\label{Equation: vlb_loss}
	\begin{align}
		L_{vlb}&=\mathbb{E}_q\{{L_T}+L_{t-1}+\cdots+L_0\}\\
        L_T:&=\text{D}_{\rm{KL}}({q(x_T|x_0)}\parallel{p(x_T)})\\
        L_{t-1}:&=\text{D}_{\rm{KL}}({q(x_{t-1}|x_t,x_0)}\parallel{p_\theta(x_{t-1}|x_t})), \ t>1\\
        L_0:&=-{\rm{log}}{p_\theta({x_0}|{x_1})}
	\end{align}
\end{subequations}
where $\text{D}_{\rm{KL}}(p \parallel q)$ represent the Kullback–Leibler divergence between two distributions $p$ and $q$. In Eq.~(\ref{Equation: vlb_loss}), ${q(x_{t-1}|x_t,x_0)}$ is tractable:
\begin{subequations}
\label{Equation: qxt-1}
	\begin{align}
		q(x_{t-1}|x_t,x_0)&=\mathcal{N}(x_{t-1};\tilde{\mu}_t(x_t,x_0),\tilde{\beta}_t{I})\\
        \tilde{\mu}_t(x_t,x_0)&={\frac{\sqrt{\alpha_{t-1}}\beta_t}{1-\alpha_t}x_0}+{\frac{\sqrt{1-\beta_t}(1-\alpha_{t-1})}{1-\alpha_t}x_t}\\
        \tilde{\beta}_t&=\frac{1-\alpha_{t-1}}{1-\alpha_t}\beta_t
	\end{align}
\end{subequations}
Assume that the variance ${\Sigma_\theta(x_t,t)}$ in the above denoising diffusion process is a constant-controlled diagonal matrix ${{\Sigma_\theta(x_t,t)}={\sigma_t^2}I}$, in this way, the training can be re-formulated solely to estimate ${\mu_\theta}$:
\begin{equation}
{p_\theta(x_{t-1}|x_t)=\mathcal{N}(x_{t-1};{\mu_\theta(x_t,t)},{\sigma_t^2}I})
\end{equation}
Then, ${L_{t-1}}$ can be re-written as: 
\begin{equation}
\label{Equation: lt-1}
L_{t-1}=\mathbb{E}_q\{\frac{1}{2{\sigma_t^2}}\parallel{\tilde{\mu}_t(x_t,x_0)-\mu_\theta(x_t,t)}\parallel_2^2\}+C
\end{equation}
where $C$ is a constant that does not participate in the optimization process and can be omitted.

By introducing a noise prediction network ${\epsilon_\theta}$, combined with Eq.~(\ref{Equation: xt}) and Eq.~(\ref{Equation: qxt-1}), ${\mu_\theta}$ can be expressed as:
\begin{equation}
\label{Equation: mu_theta}
\mu_\theta(x_t,t)=\frac{1}{1-\beta_t}(x_t-\frac{\beta_t}{\sqrt{1-\alpha_t}}\epsilon_\theta(x_t,t))
\end{equation}
Therefore, by setting ${{\sigma_t^2}=\beta_t}$, based on Eq.~(\ref{Equation: lt-1}) and Eq.~(\ref{Equation: mu_theta}), we can finally obtain the training objective function as:
\begin{equation}
\label{Equation: vlb_simple}
L_{t-1}=\mathbb{E}_{x_0,\epsilon}\{\frac{1}{\lambda_t}\parallel{\epsilon-\epsilon_\theta({\sqrt{\alpha_t}x_0}+{\sqrt{1-\alpha_t}\epsilon},t)}\parallel^2\}
\end{equation}
where $\lambda_t=(1-\beta_t)(1-\alpha_t)/\beta_t$ is a time-variant scaling factor on different loss terms through different denoising steps.

\subsection{The MetaEarth Self-Cascaded Generation Framework}

We propose a resolution-guided self-cascading framework for generating various scenes and resolutions. Fig.~\ref{fig:method_overview} shows the overall structure of the proposed MetaEarth. This framework enables recursive enhancement of image resolution using a unified generation model. 

The entire generation process consists of a series of stages, progressing from low-resolution images to high-resolution images. In each generation stage, the model is conditioned on the previously generated low-resolution images and their corresponding spatial resolutions. The low-resolution images provide scene categories and semantic information, while spatial resolution information enables the model to perceive and represent features of images at different scales. 

Suppose $x_0^{(k)}$ and $s^{(k)}$ represent the clean image and the spatial resolution (m/pixel) at the $k$-th generation stage. $x_0^{(k)}$ and $s^{(k)}$ are embedded as conditional variables when generating images at $(k+1)$-th stage. Assuming that at each generation stage, the model increases the resolution of the input image by a factor of $N$, e.g. $N=4$, if the image generated at stage $k$ has a size of $(H \times W)$ pixels, then at stage $k+1$, we will obtain a high-resolution image with a size of $(NH \times NW)$.

Assuming there are no memory constraints, by repeating the aforementioned operation $m$ times based on stage $k$, we will ultimately obtain a series of images with varying resolutions and pixel sizes: 
\begin{equation}
    \mathcal{X} = \{ x_0^{(k)}, x_0^{(k+1)}, \cdots, x_0^{(k+m)}  \}
\end{equation}
where the image $x_0^{(k+m)}$ generated from $(k+m)$-th stage will eventually has a size of $(N^mH\times N^mW)$ pixels. To achieve the above process with reasonable memory and computational costs, we designed a sliding window generation process and a noise sampling strategy that enables the generation of continuous, unbounded scenes in a memory-efficient manner. These approaches will be detailed in Subsection~\ref{sec:unbounded}.

In each generation stage, both the previously generated image and the time-resolution embeddings are utilized as conditional input variables. For the conditional input $x_0^{(k)}$, whose dimensions mismatched with $x_t^{(k+1)}$, inspired by~\cite{sr3} and~\cite{srdiff}, we redesigned their network architecture. We begin by encoding the low-resolution image $x_0^{(k)}$ with an encoder $E_{lr}$. Subsequently, we use a set of upsampling and convolution layers $F_{up}$ to align the dimensions of the feature maps with $x_t^{(k+1)}$. Lastly, we merge the features by concatenating $x_t^{(k+1)}$ with the feature maps along the channel dimension.
\begin{equation}
\label{Equation: lr_encoder}
\tilde{x}_t^{(k+1)}=\text{cat}[x_t^{(k+1)},F_{up}(E_{lr}(x_0^{(k)}))]
\end{equation}
where $\text{cat}(\cdot)$ represents the concatenation operation between two tensors along their channel dimension.

For the spatial resolution $s$, we firstly encode it through a frequency encoding transformation:
\begin{equation}\label{Equation: sin-cos}
\begin{split}
f_\omega(s^{(k)})=[&\cos(s^{(k)}f_1),\sin(s^{(k)}f_1),\cos(s^{(k)}f_2),\sin(s^{(k)}f_2),\\
    &\cdots,\cos(s^{(k)}f_n),\sin(s^{(k)}f_n)]    
\end{split}
\end{equation}
where the frequency factors are defined as:
\begin{equation}
    f_i=\exp(-\frac{i}{n}\log(\omega))=\omega^{-\frac{i}{n}}, i=1,2,\cdots,n
\end{equation}
To let the model perceive more subtle resolution changes, we set ${\omega}$ as a large number, $\omega=10^4$. Then, the encoded resolution is further projected by a trainable Multi-Layer Perception network $F_{mlp}$.The complete encoding process for ${s}$ can be represented as follows:
\begin{equation}
\label{Equation: sr_encoder}
e_s^{(k)}=F_{mlp}(f_\omega(s^{(k)}), \theta_s)
\end{equation}
where $\theta_s$ are the learnable parameters of the $F_{mlp}$.

Through the above process, we obtain the spatial resolution embedding at $k$-th generation stage $e_s^{(k)}\in{\mathbb{R}^D}$.
Similarly, the time-step variables $t\in\{1,\dots,T\}$ during the denoising process can be also encoded through the frequency encoding transformation $f_\omega(t)$ and then projected into a $D$-dimensional embedding vector with an additional MLP:
\begin{equation}
\label{Equation: timestep_encoder}
e_t=F_{mlp}(f_\omega(t), \theta_t)
\end{equation}
where $\theta_t$ are the learnable parameters.
We finally add $e_s^{(k)}$ and $e_t$ together to produce the final conditional embedding vector at the $t$-th denoising step from the $k$-th generation stage:
\begin{equation}
e_t^{(k)} = e_s^{(k)} + e_t
\end{equation}

Given the conditional embedding vector $e_t^{(k)}$ all together with the conditional image features $\tilde{x}_t^{(k)}$, we uniformly express them as the condition variables of the $k$-th generation stage and $t$-th denoising step:
\begin{equation}
    c_t^{(k)} = \{e_t^{(k)}, \tilde{x}_t^{(k)}\}
\end{equation}

Based on Eq.~(\ref{Equation: reverse process}) while maintaining the forward process unchanged, we incorporate the condition $c_t^{(k)}$ as input to build a conditional denoising model:
\begin{subequations}
\label{Equation: conditional diffusion}
	\begin{align}
	p_\theta(x_{0:T}|c_t^{(k)})&=p(x_T)\prod_{t=1}^{T}{p_\theta(x_{t-1}|x_t, c_t^{(k)}}) \\
        p_\theta(x_{t-1}|{x_t},c_t^{(k)})&=\mathcal{N}(x_{t-1};\mu_\theta(x_t,t,c_t^{(k)}),\sigma_t^2{I})
	\end{align}
\end{subequations}

Through the aforementioned design, we can finally generate images of different resolutions in a self-cascading manner driven by resolution and time-step encodings.

\subsection{Unbounded Image Generation}\label{sec:unbounded}

\begin{figure}[t]
	\centering
	\includegraphics[width=\linewidth]{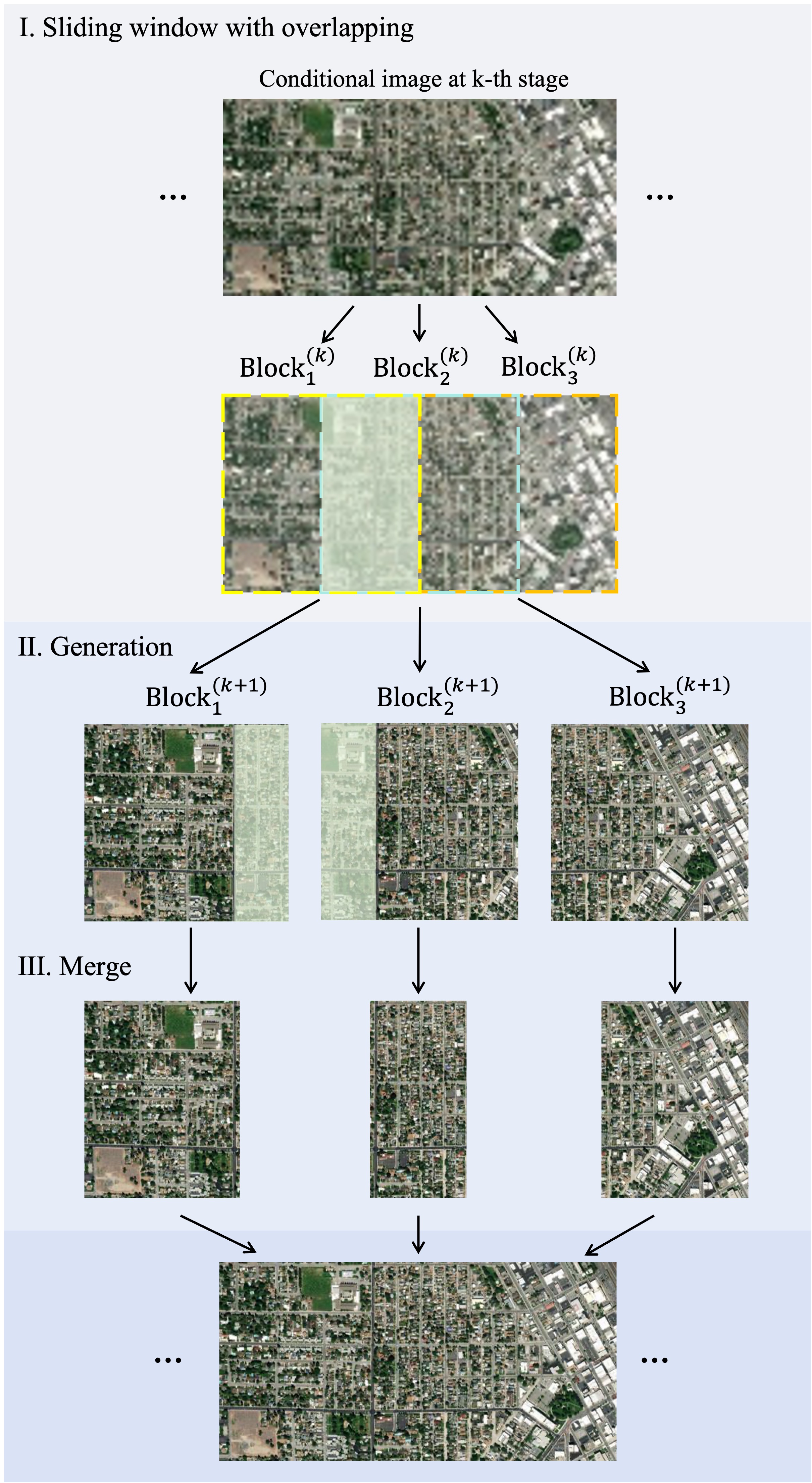}	
	\caption{To generate unbounded images, firstly, the generated image from the previous stage is cropped into overlapping image blocks as conditions. Then, with the proposed noise sampling strategy, the shared regions between adjacent image blocks generate similar content. Lastly, the generated images are tiled and re-organized.}
	\label{fig:method_sliding_windows}
\end{figure}

To achieve the generation of large-scale remote sensing images of arbitrary sizes, we propose an unbounded image generation method including a memory-efficient sliding window generation pipeline and a noise sampling strategy. Fig.~\ref{fig:method_sliding_windows} illustrates the proposed method.

In the self-cascading generation framework, we crop the conditional input images into a set of image blocks to control the memory overhead within an acceptable range. However, simple cropping and stitching will result in noticeable seams at junctions. We analyze the occurrence of seams from two factors in generative diffusion models: the conditions and the noise sampling. 

From the perspective of generation conditions, when the receptive field of the model extends beyond the image boundaries, the generated content may differ semantically from adjacent images. To address this issue, we consider blocks as sliding windows with overlapping of 1/2 window area so that pixels originally located at the boundaries are repositioned as central areas. The overlapped regions serve as semantic transition blocks between adjacent images so that the semantic discontinuity between image blocks during stitching can be solved.

Additionally, the initial noise sampling is another crucial factor affecting the continuity between tiles. Due to the randomness of the noise sampling, the generated content within the overlapping regions may vary, resulting in miss-alignment of between adjacent blocks. To mitigate pixel-level discrepancies, we propose a noise sampling strategy that, without modifying the model training process, generates identical or near-identical content in the overlapping regions, ensuring pixel-level continuity. 

During the reverse process of conditional diffusion, we adopt the sampling method of DDIM~\cite{ddim}. The denoising sampling process from $x_t^{(k)}$ to $x_{t-1}^{(k)}$ is represented as follows:
\begin{equation}
\label{Equation: ddim}
\begin{split}
x_{t-1}^{(k)}=\sqrt{\alpha_{t-1}}(\frac{{x_t}^{(k)}-\sqrt{1-\alpha_t}\epsilon_\theta(x_t^{(k)}, c_t^{(k)})}{\sqrt{\alpha_t}})\\
+\sqrt{1-\alpha_{t-1}-{\sigma_t^2}}\cdot{\epsilon_\theta(x_t^{(k)},c_t^{(k)})}+{\sigma_t}{\epsilon_t}
\end{split}
\end{equation}
where ${{\epsilon_t}\sim{\mathcal{N}(0,I)}}$, and ${{\sigma_t}={\eta}\sqrt{(1-\alpha_{t-1})/(1-\alpha_t)}\sqrt{1-{\alpha_t}/\alpha_{t-1}}}$. ${\eta}$ is a predetermined factor ranging from 0 to 1. 
When ${\eta=1}$ the DDIM sampling process is identical to DDPM. When ${\eta=0}$, the denoising process is fully determined, and the generated image is solely determined by the initial noise and the condition $c_T^{(k)}$.
In our method, we set ${\eta=0}$, that is, model it as a completely deterministic sampling process. In this setting, since $x_0^{(k)}$ is solely determined by the initial noise ${x_T}$ and the condition $c_T^{(k)}$, we represent the generation process using an injective function:
\begin{align}
\label{Equation: f(n,c)}
x_0^{(k)} = f_\theta(x_T, c_T^{(k)})
\end{align}

Therefore, if the receptive field of the denoising network is smaller than the overlapped region, with the same condition $c_T^{(k)}$ and the same initial noise at the overlapping regions, we can easily prove that the generated images at the center of the overlapped region will be strictly same.

\begin{figure}[t]
	\centering
	\includegraphics[width=0.6\linewidth]{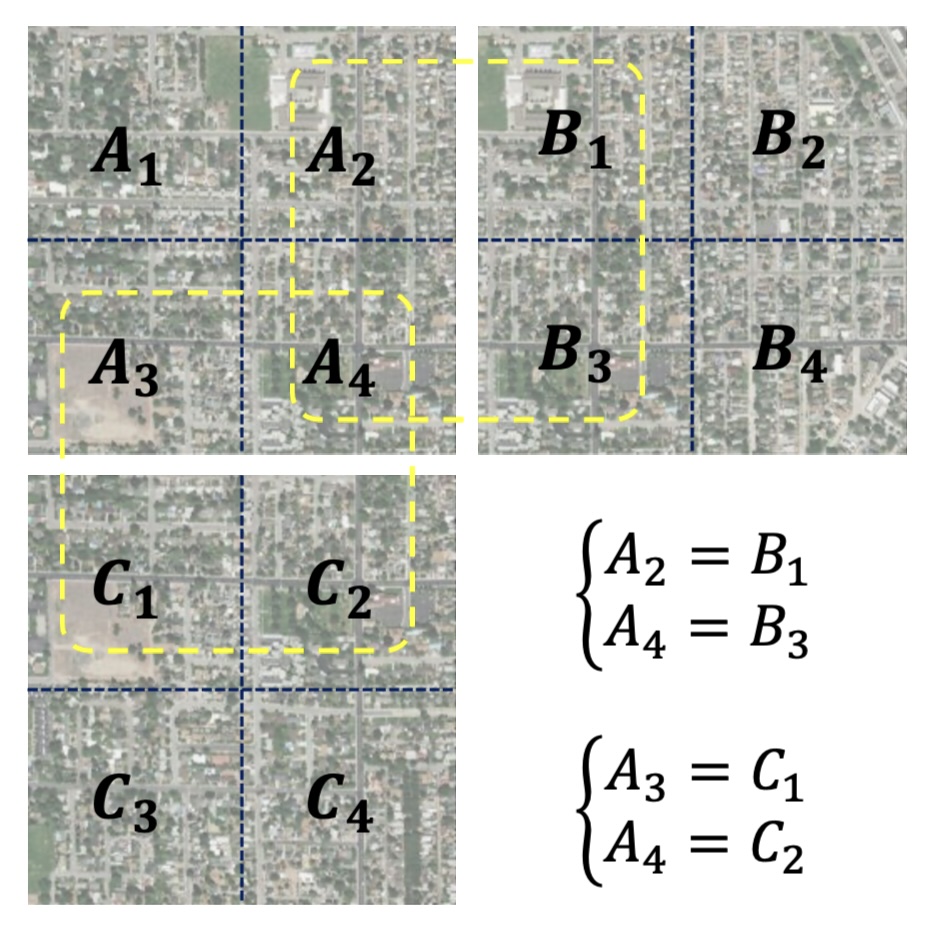}	
	\caption{Necessary conditions on initial noise sampling for generating continuous unbounded scenes.}
	\label{fig:method_noise_sampling}
\end{figure} 

Based on the above analysis, as shown in Fig.~\ref{fig:method_noise_sampling}, to ensure that vertically adjacent image blocks generate similar content in their overlapping regions, we need their initial Gaussian noise to satisfy the conditions ${A_3=C_1}$ and ${A_4=C_2}$. Similarly, to ensure that horizontally adjacent image blocks generate similar content in their overlapping regions, we need ${A_2=B_1}$ and ${A_4=B_3}$. To make sure that all the above conditions are always true during the generation process, we can first sample for one block, and then make copies for the corresponding neighboring blocks. 
To further simplify the sampling process, in our implementation, we set all blocks with the same initial noise, i.e., ${N_1=N_2=N_3=N_4}$, where $N=\{A, B, C\}$.

\subsection{Implementation Details}

\subsubsection{Denoising Network Design}

For the dataset constructed above, we set the model to increase the input image resolution by a factor of $N=4$. For the noise schedule design, we adopt a linear schedule ranging from a minimum of 0.0015 to a maximum of 0.0155. During training, the number of sampling steps is set to 1000, while during inference, we utilize the DDIM acceleration strategy with the number of sampling steps set to 50.

We design a U-Net-like architecture to predict noise, with a total number of about 600 million parameters. The encoder and decoder of the network are composed of 5 blocks each. $\tilde{x}_t^{(k+1)}$ undergoes 4 rounds of 2$\times$ downsampling/upsampling. The number of channels in each block is multiplied by a factor of [1,2,4,8,8] based on the base number of channels. Each block contains 3 ResBlocks, with AttentionBlocks included specifically for blocks with an 8-fold channel multiplier. We utilized RRDBNet~\cite{esrgan} as the encoder for conditional images. 

For each feature map $h$ within every ResBlock, we divide the embedding $e_t^{(k)}$ into two parts using the chunk function in PyTorch and then scale the feature map $h$ with the chunked embeddings:
\begin{equation}
\label{Equation: scale_shift}
\begin{split}
&e_{\text{scale}}, e_{\text{shift}}={\texttt{chunk}}(e_t^{(k)},2)\\
&h \leftarrow {(1+e_{\text{scale}})}\cdot h + e_{\text{shift}}
\end{split}
\end{equation}
With the above design, we can integrate both low-resolution input and spatial resolution to predict the noise ${\epsilon_\theta}$ for each timestep.

\subsubsection{Training Details}

When training a single model for cascaded generation, two main challenges arise: Firstly, high-resolution and low-resolution images are from different sensors and are captured under varying spatiotemporal conditions. This results in significant differences in style, content, and details between pairs of high and low-resolution images, leading to potential mismatches between image pairs. Secondly, during the iterative inference process of self-cascading generation, each input image to the generation model in the cascading steps originates from the model's previous output. However, the input data utilized during model training may differ in distribution from the input data during inference. This distributional shift may result in distorted or unreasonable image generation results. To address the aforementioned issues, inspired by~\cite{realesrgan}, we introduce high-order degradation to simulate the above factors. We artificially degrade high-resolution images into low-resolution ones, thus constructing training image pairs. The high-order degradation process can be represented as follows:

\begin{equation}
\label{Equation: high-order degration}
x_{lr}=\mathcal{D}^{n}(x_{hr})=({\mathcal{D}_1}\circ{\mathcal{D}_2}\circ{\cdots}\circ{\mathcal{D}_n})(x_{hr})
\end{equation}

where ${\mathcal{D}_i}$ represents a simple degradation model typically comprised of blur, scaling, noise addition, and JPEG compression, which can be expressed as:

\begin{equation}
\label{Equation: simple degration}
x_{lr}={D_i}(x_{hr})=[(x_{hr}{\ast}{k}){\downarrow}_{r}+n]_{\rm{JPEG}}
\end{equation}

where ${k}$, ${r}$, ${n}$, ${[\cdot]_{JPEG}}$ represent the blur kernel, the scaling factor, noise, JPEG compression, respectively. During training, We directly apply the high-order degradation, composed of two simple degradation models, in the pixel domain of high-resolution images to downsample real images into corresponding low-resolution ones, i.e. $n=2$:

\begin{equation}
\label{Equation: used degration}
x_{lr}=\mathcal{D}^{2}(x_{hr})=({\mathcal{D}_1}\circ{\mathcal{D}_2})(x_{hr})
\end{equation}

Following the parameter settings as described in~\cite{realesrgan}, in terms of blur kernels, we chose isotropic and anisotropic Gaussian blur kernels, as well as sinc filters. During the scaling phase, we randomly used nearest, bilinear, and bicubic interpolation methods to enlarge or downscale the images; for noise, we randomly added Gaussian or Poisson noise to the images, with the noise type randomly selected as colour or grey; for the JPEG compression process, we randomly compressed the images with quality factors ranging from 30 to 95.

During inference, we do not apply any degradation - the low-resolution conditional input can be either real images or images generated from the previous cascading steps.

When constructing the final loss function for denoising model training, we introduced the Perception Prioritized (P2) weighting and imposed a weighting factor on different denoising steps~\cite{p2weighting}. This design forces the model to focus more on restoring perceptually rich content during training, resulting in higher-quality image generation. Based on Eq.~(\ref{Equation: vlb_simple}), the final loss function for training the denoising network is defined as follows:

\begin{equation}
\label{Equation: loss function}
L(\theta)={\sum_{t=1}^T}{\lambda_t^{\prime}}{L_{t-1}}
\end{equation}

where the weighting factors $\lambda_t^{\prime}$ are defined as:

\begin{equation}
\label{Equation: lambda_t'}
{\lambda_t^{\prime}}=\frac{\lambda_t}{(k+\text{SNR}(x_t))^{\gamma}}
\end{equation}

In this equation, $k$ and $\gamma$ are hyperparameters that were both set to 1 in the experiments, following the suggestions in ~\cite{p2weighting}. The term $\text{SNR}(x_t)$ represents the signal-to-noise ratio of the noisy image.

Our model is implemented using PyTorch. We utilize the AdamW optimizer with an initial learning rate of 2e-6 to optimize parameters and set the total number of epochs to 30. To increase the training batch size, we employ gradient accumulation and multi-GPU parallelism. The batch size per GPU is set to 1, and gradient accumulation is set to 8. The training process takes more than 2,000 GPU hours on a set of NVIDIA RTX4090 GPUs. The training is initialized from scratch without loading pre-trained parameters.

\section{EXPERIMENTS}
\subsection{Experimental Setup}
\subsubsection{Dataset}

To generate multi-resolution remote sensing images for any region globally, we collected a large dataset, which includes three levels of 256$\times$256 sized remote sensing images at arbitrary latitudes and longitudes from Google Earth, corresponding to resolutions of 64m/pix, 16m/pix, and 4m/pix. These data encompass nearly all of the Earth's geographical and environmental conditions, including cities, forests, deserts, oceans, glaciers, and more. Then, we cleansed the dataset, removing some ocean images that were highly repetitive, as there was no need to use an excessive amount of similar data for model training. We also eliminated images that contained noise or significant cloud cover. Lastly, for the handpicked dataset, we randomly sampled approximately 1,000,000 non-overlapping images at each resolution (a total of about 3,100,000 images) for model training, and additionally randomly sampled about 140,000 images, which were divided into a validation set and a test set in a 1:1 ratio. The training set images at various resolutions are shown in Table \ref{tab:dataset}.

\begin{table}[!ht] 
\renewcommand{\arraystretch}{1.3}
\captionsetup{justification=centering, labelsep=newline, font=small}
\caption{The number of training data at various resolutions.}
\label{tab:dataset}
\centering
\begin{tabular}{c|c}
	\toprule
	{Spatial Resolution} & {Image Count} \\
	\midrule
        64m/pixel & 916,509\\
	16m/pixel & 1,022,194\\
        4m/pixel & 1,186,312\\
	\bottomrule
\end{tabular}
\end{table}

We strictly adhered to Google Earth's copyright and guidelines during our data collection process, and upon acceptance of the paper, we deleted all the training data.

\subsubsection{Evaluation Metrics}

We use Fréchet Inception Distance (FID)~\cite{fid} to evaluate the quality and diversity of the generated images. A lower FID indicates higher fidelity and diversity in the generated images. FID quantifies the similarity between the feature representations of generated and real images using a pre-trained deep convolutional neural network. The FID score is defined as follows:
\begin{equation}
\label{Equation: fid}
\rm{FID}=\parallel{\mu_{r}-\mu_{g}}\parallel_2^2+{\mathrm{Tr}}(\Sigma_r+\Sigma_g-2({\Sigma_r}{\Sigma_g})^{\frac{1}{2}})
\end{equation}
where ${\mu_{r}}$ and ${\mu_{g}}$ are the mean feature vector of the real and generated images, respectively. ${\Sigma_r}$ and ${\Sigma_g}$ are the covariance matrices of the feature representations of the real and generated images, respectively. The features used for computing the FID are extracted from the final average pooling layer of a pre-trained Inception v3 network~\cite{szegedy2016rethinking}.

\subsection{Qualitative Analysis}

\subsubsection{Global-Scale Image Generation}

\begin{figure*}
	\centering
    \captionsetup{justification=raggedright, singlelinecheck=false} 
	\includegraphics[width=0.98\linewidth]{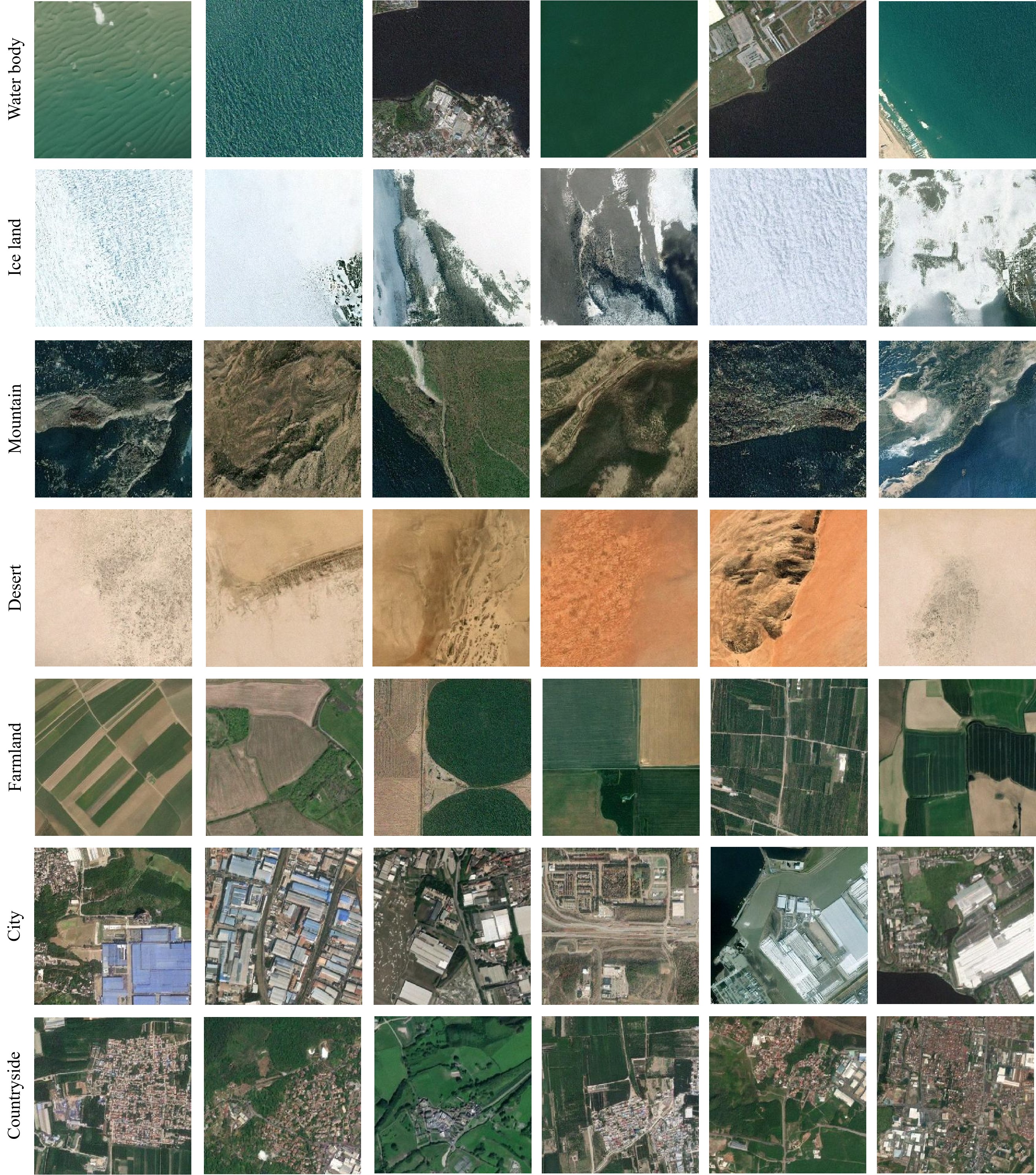}	
	\caption{Images of various land features across the globe generated by MetaEarth, including water bodies, mountains, deserts, farmland, cities, and countryside areas.}
	\label{fig:exp_world_gallary}
\end{figure*} 

\begin{figure*}
	\centering
    \captionsetup{justification=raggedright, singlelinecheck=false} 
	\includegraphics[width=0.9\linewidth]{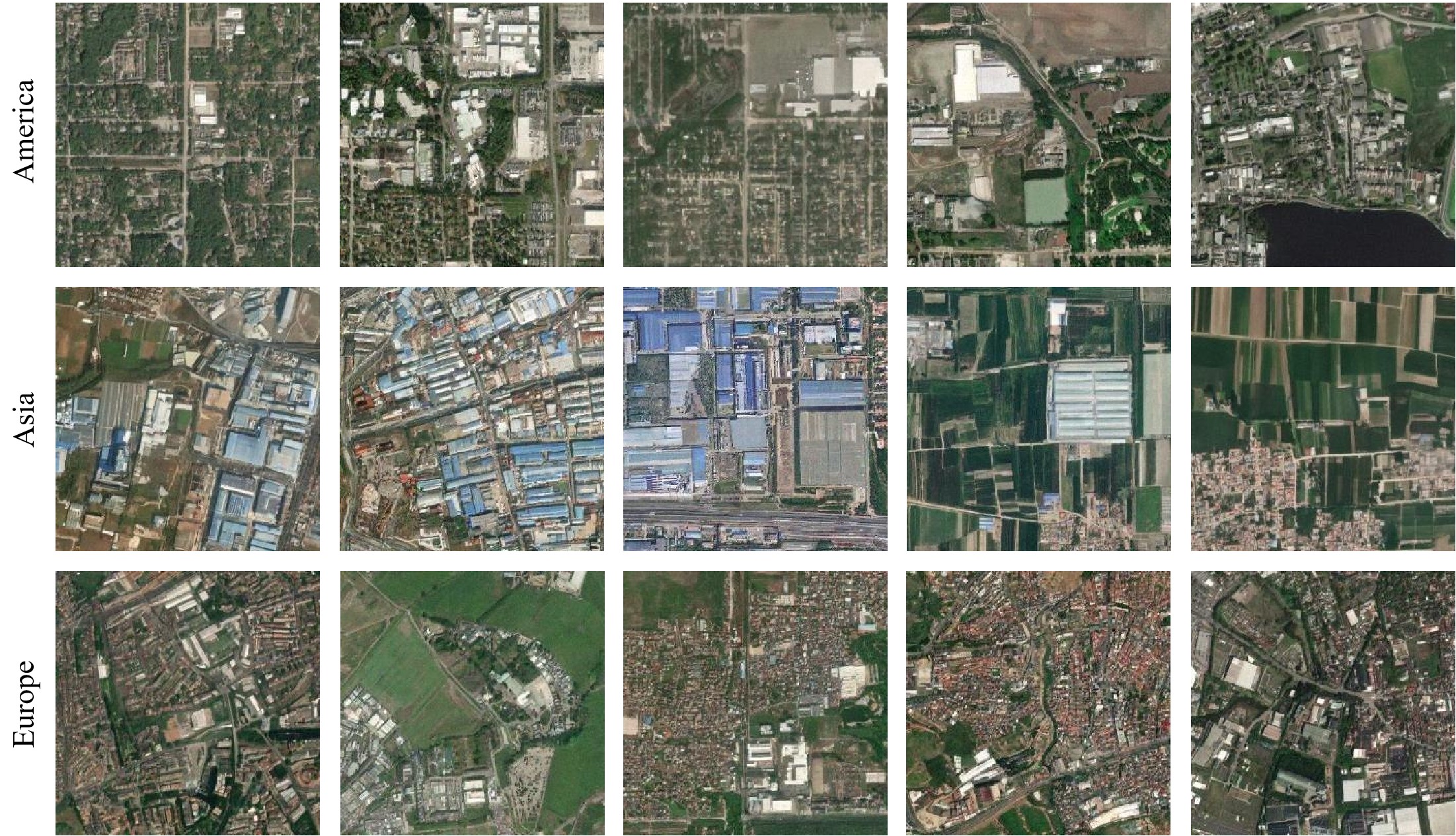}	
	\caption{MetaEarth-generated city images with different continental styles. We use large-scale low-resolution images of 256m/pixel in the Americas, Europe, and Asia as condition guidance, and the model can generate urban images consistent with the regional style.}
	\label{fig:exp_city_compare}
\end{figure*}

Our experiments show that the proposed MetaEarth can generate a variety of remote sensing scenes worldwide, including glaciers, snowfields, deserts, forests, beaches, farmlands, industrial areas, residential areas, etc. Fig.~\ref{fig:teaser} and Fig.~\ref{fig:exp_world_gallary} showcase some results of our model. Owing to the guidance from the conditional input, the model is capable of generating vivid and diverse images with distinct regional characteristics at different resolutions. We also evaluated the semantical guidance provided by the image conditional input. In Fig.~\ref{fig:exp_city_compare}, we use large-scale low-resolution images of 256m/pixel in the Americas, Europe, and Asia as image condition guidance, and we can see the model can generate urban images consistent with the regional style.

\subsubsection{Multi-Resolution Image Generation}

\begin{figure}
	\centering
	\includegraphics[width=\linewidth]{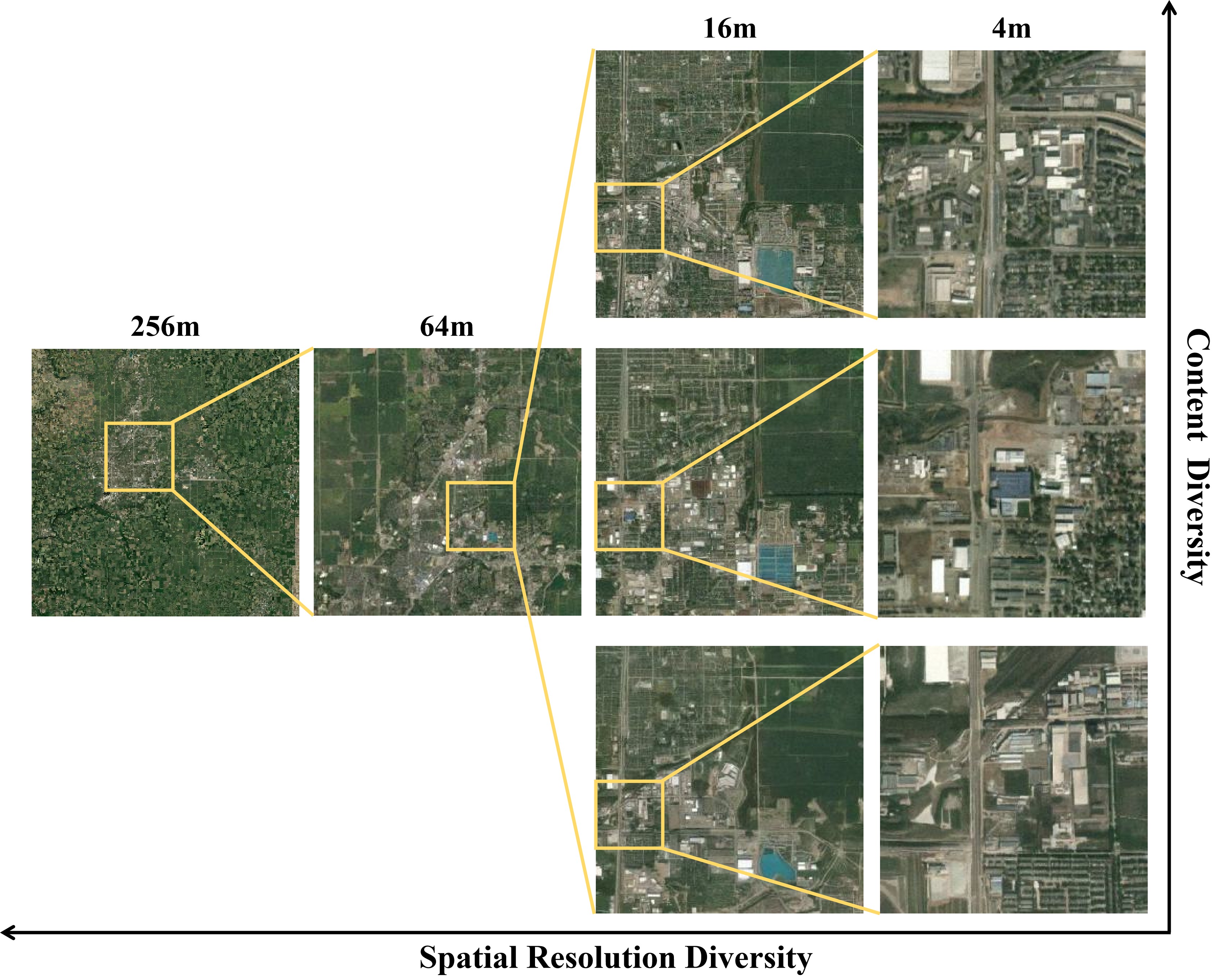}	
	\caption{Visualization of the generated diverse scenes. By combining the proposed self-cascading generation framework with different noise sampling conditions, we can generate diverse and ``parallel'' scenes under the same initial conditions.}
	\label{fig:exp_diversity}
\end{figure} 

The self-cascading generation framework in our method enables the model to generate images with both spatial resolution diversity and content diversity. Our experiments show that using high-order degradation models to downsample high-resolution images for training pairs improved the model's generalization ability. Consequently, the model can take generated images as input to produce higher-resolution images with richer details. Fig.~\ref{fig:exp_diversity} illustrates the generated results of multiple resolutions. At the very beginning stage, the model is conditioned on images with a resolution of 256 meters, finally resulting in generated images with a spatial resolution of 4 meters.
The results also suggest the model's strong diversity-generating capabilities.  We can see that as the number of generation stages increases, the content diversity between images gradually increases due to the variability introduced by the generation model.

\subsubsection{Arbitrary-sized Image Generation}

\begin{figure*}[t]
	\centering
	\includegraphics[width=0.95\linewidth]{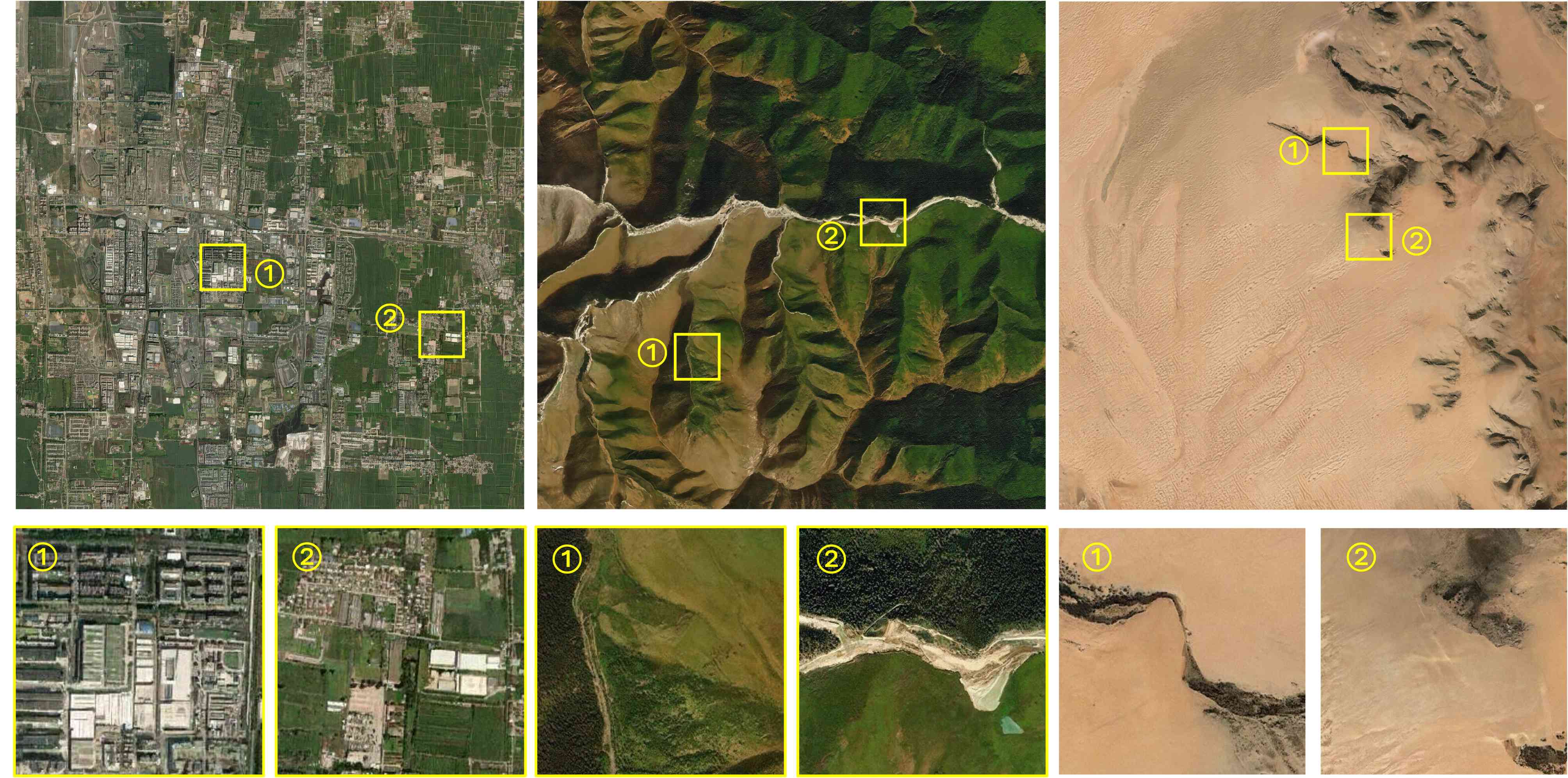}
    \captionsetup{justification=raggedright, singlelinecheck=false}
    \caption{Some examples of high-resolution large-sized images generated by our model.}
	\label{fig:exp_4k}
\end{figure*}

\begin{table*}[t]
\renewcommand{\arraystretch}{1.3}
\captionsetup{justification=centering, labelsep=newline, font=small}
\caption{Ablation studies on the spatial resolution guidance. We divided the Earth into different regions and assessed the quality (FID, a lower score indicates better) of images generated in each region. ${lng}$ denotes the longitude.}
\label{tab:sr-ablation}
\centering
\begin{tabular}{c|c|ccccccc}
	\toprule[1pt]
         ~ & Settings & ${-180 \leq lng<-90}$ & ${-90 \leq lng<0}$ & ${0 \leq lng<90}$ & ${90 \leq lng<180}$ & {Northern Hemisphere} & {Southern Hemisphere} & {Total}\\
    \midrule
    \multirow{2}{*}{64m} & {w/o sr} & 33.23 & 40.81 & 42.29 & 45.10 & 29.50 & ${\boldsymbol{42.11}}$ & 27.79   \\
          ~ & {w/ sr} & ${\boldsymbol{30.32}}$ & ${\boldsymbol{37.90}}$ & ${\boldsymbol{39.04}}$ &  ${\boldsymbol{42.16}}$ & ${\boldsymbol{25.01}}$ & 44.16 & ${\boldsymbol{23.97}}$\\
    \midrule
    \multirow{2}{*}{16m}  & {w/o sr} & 19.72 & 21.12 & 26.03 & 26.58 & 16.84 & ${\boldsymbol{25.50}}$ & 16.00 \\
          ~ & {w/ sr} & ${\boldsymbol{19.36}}$ & ${\boldsymbol{21.91}}$ & ${\boldsymbol{24.65}}$ & ${\boldsymbol{25.47}}$ & ${\boldsymbol{15.76}}$ & 27.76 & ${\boldsymbol{15.12}}$\\
    \midrule
    \multirow{2}{*}{4m}  & {w/o sr} & 19.47 & 13.91 & 18.65 & 17.47 & 13.93 & ${\boldsymbol{22.02}}$ & 13.43 \\
          ~ & {w/ sr} & ${\boldsymbol{18.79}}$ & ${\boldsymbol{15.67}}$ & ${\boldsymbol{18.13}}$ & ${\boldsymbol{17.26}}$ & ${\boldsymbol{13.68}}$ & 25.63 & ${\boldsymbol{13.32}}$\\
    \midrule
    \multirow{2}{*}{4m*} & {w/o sr} & 141.15 & 131.65 & 119.67 & 111.16 & 116.63 & 147.16 & 92.61 \\
          ~ & {w/ sr} & ${\boldsymbol{134.92}}$ & ${\boldsymbol{129.60}}$ & ${\boldsymbol{113.83}}$ & ${\boldsymbol{107.67}}$ & ${\boldsymbol{112.42}}$ & ${\boldsymbol{139.63}}$ & ${\boldsymbol{84.75}}$\\
    \bottomrule
\end{tabular}
\end{table*}

Fig.~\ref{fig:exp_4k} shows some examples of high-resolution large-scale images generated with the proposed sliding window generation. Although we applied block-by-block generation, it can be observed that the proposed method generates continuous land features across image blocks. This indicates that our window overlapping and noise sampling strategy greatly alleviate visual discontinuities caused by image block stitching, thereby achieving boundless and arbitrary-sized image generation.

\subsubsection{Generalization to Unseen Conditions}

\begin{figure*}
	\centering
	\includegraphics[width=0.95\linewidth]{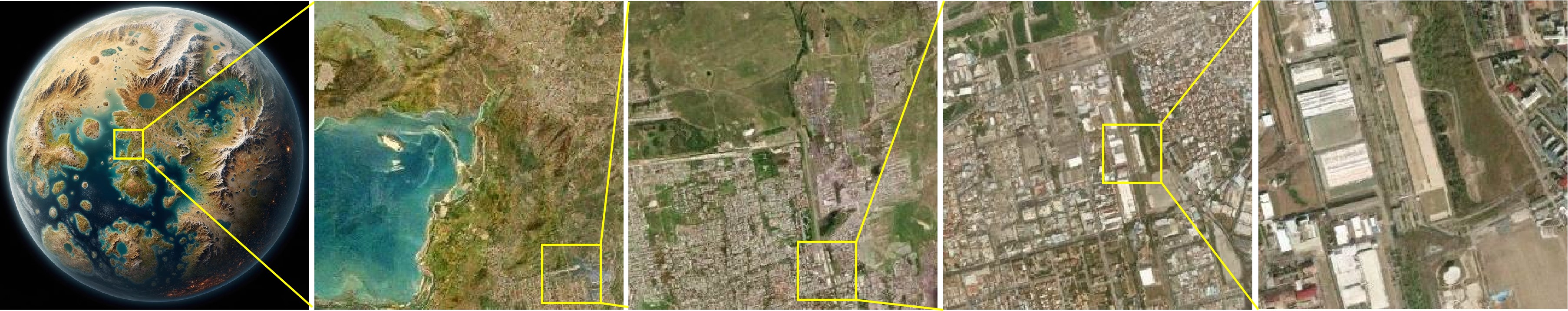}
    \captionsetup{justification=raggedright, singlelinecheck=false}
    \caption{The self-cascaded generation results from MetaEarth with unseen conditions.}
	\label{fig:exp_pandora}
\end{figure*} 

\begin{figure}
	\centering
	\includegraphics[width=\linewidth]{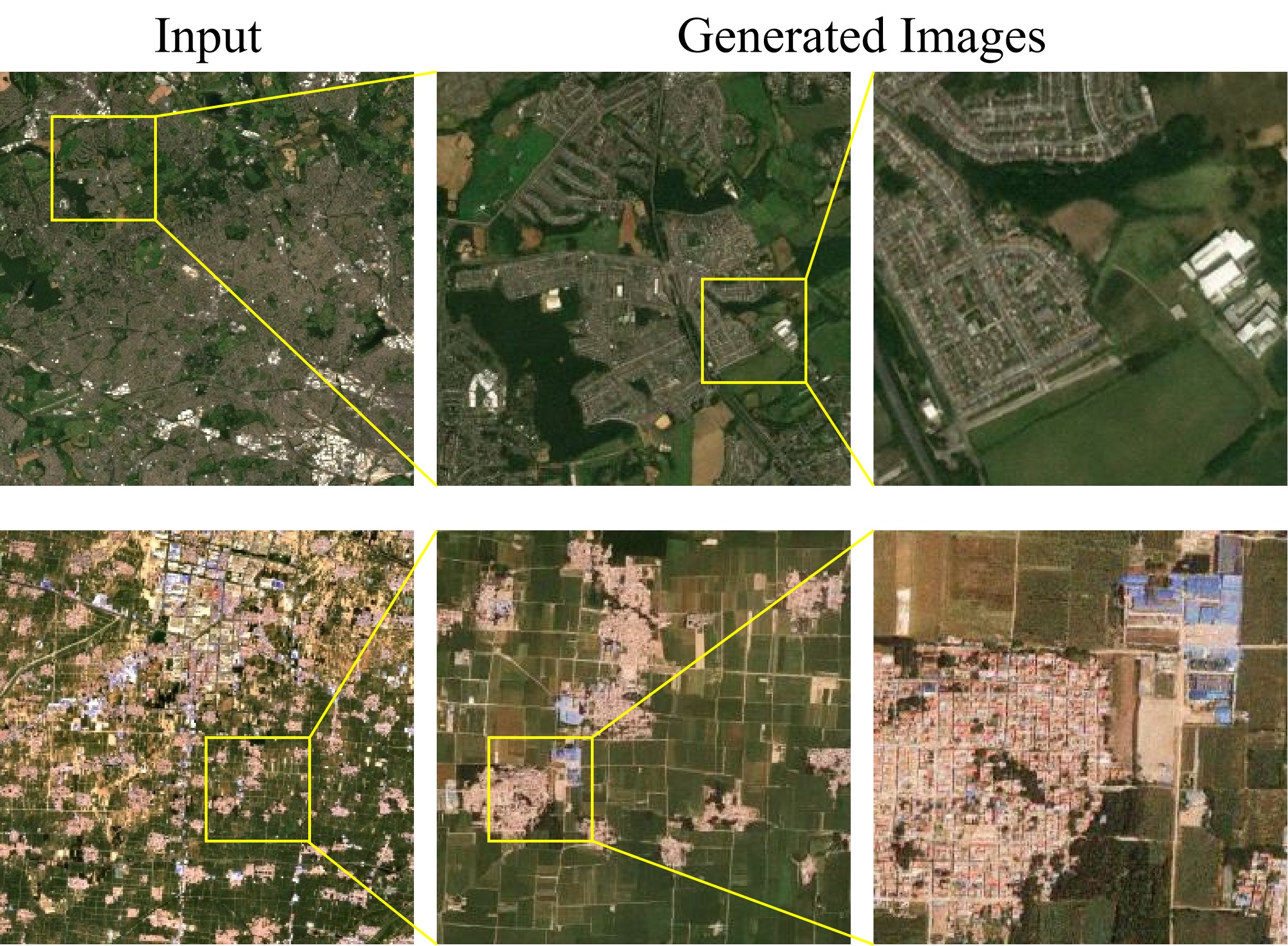}	
	\caption{The self-cascaded generation results from MetaEarth with image sourced from Sentinel-2 satellite.}
	\label{fig:exp_sentinel}
\end{figure} 

Thanks to being trained on large-scale data, our model possesses strong generalization capabilities and performs well even on unseen scenes. We conducted two sets of experiments to verify the generalization capability of our MetaEarth model on unseen conditions. In the first experiment, we aimed to create a ``parallel world'' and used a low-resolution map of ``Pandora Planet'' (generated by GPT-4V) as the initial condition for our model and then generated higher-resolution images sequentially. As shown in Fig.~\ref{fig:exp_pandora}, MetaEarth is capable of generating images that have a reasonable land cover distribution along with realistic details. In the second experiment, we utilized real remote sensing images collected by the Sentinel-2 satellite as low-resolution inputs to generate multi-resolution images, with the results depicted in Fig.~\ref{fig:exp_sentinel}. The generated images preserved the imaging style of the original sensor while exhibiting distinct regional characteristics at various resolutions. It is noted that despite our training data not covering scenes in these two experiments, MetaEarth is still able to generate clear, semantically reasonable, and regionally styled realistic images, which indicates that our model possesses strong generation capabilities and performs well even on unseen scenes.

\subsubsection{Comparison with other text-to-image models}

\begin{figure}
	\centering
	\includegraphics[width=0.9\linewidth]{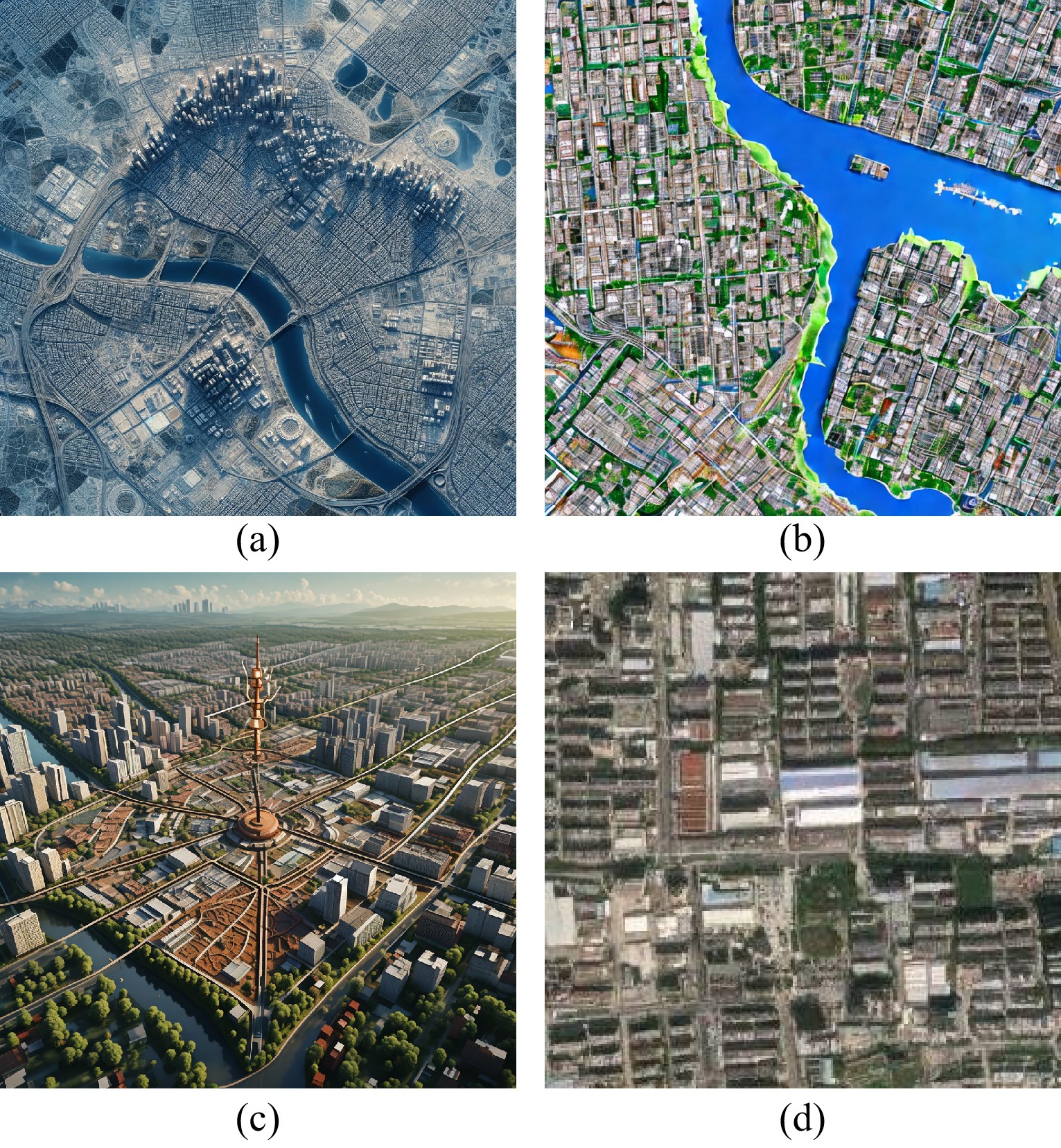}
    \captionsetup{justification=raggedright}
    \caption{Comparision between MetaEarth to other text-to-image models including GPT-4V (a), the latest version of Stable Diffusion (b), and Ernie (c). For the three compared methods, the text prompts are given as: ``Please generate a 4m/pixel-resolution satellite remote sensing image of an urban scene. The image includes detailed city infrastructure such as roads, buildings, parks, and waterways with clear visibility of the urban layout and structure."}
	\label{fig:exp_gpt4v_compare}
\end{figure}

Despite the rapid advancements in generative models represented by Stable Diffusion and DALLE, their capabilities in generating overhead remote sensing scenes are limited. In Fig.~\ref{fig:exp_gpt4v_compare}, we compare MetaEarth with other state-of-the-art text-to-image models, including GPT-4V (DALLE), the latest version of Stable Diffusion, and Ernie. We can see that MetaEarth significantly outperforms these models in terms of image realism and the rationality of scene layouts. The primary reason behind this difference is the lack of relevant remote sensing training data during the training phase of previous models. In addition, the three comparison models lack sensitivity to resolution control and are difficult to accurately capture the correspondence between resolution and image content, making it difficult to generate multi-resolution images or unbounded scenes. MetaEarth, instead, excels in these aspects, demonstrating its superior capability in generating high-quality, realistic remote sensing images across various resolutions and scene layouts.

\subsection{Ablation Study}

In this section, we conduct experiments to validate the effectiveness of the proposed methods, including the self-cascaded generation framework, resolution-guided generation and unbounded generation.

\subsubsection{Necessity of Self-cascaded Generation Framework}

The self-cascaded generation framework we propose aims to generate multi-resolution images in a stage-by-stage manner. In addition, we have explored an alternative one that is fundamentally different from our self-cascaded method, which involves generating the highest-resolution image by splicing directly and then downsampling it level-by-level to get each low-resolution image. This method contrasts with the self-cascaded generation method presented in this paper, representing two distinct framework for image generation.

\begin{figure*}
	\centering
	\includegraphics[width=0.95\linewidth]{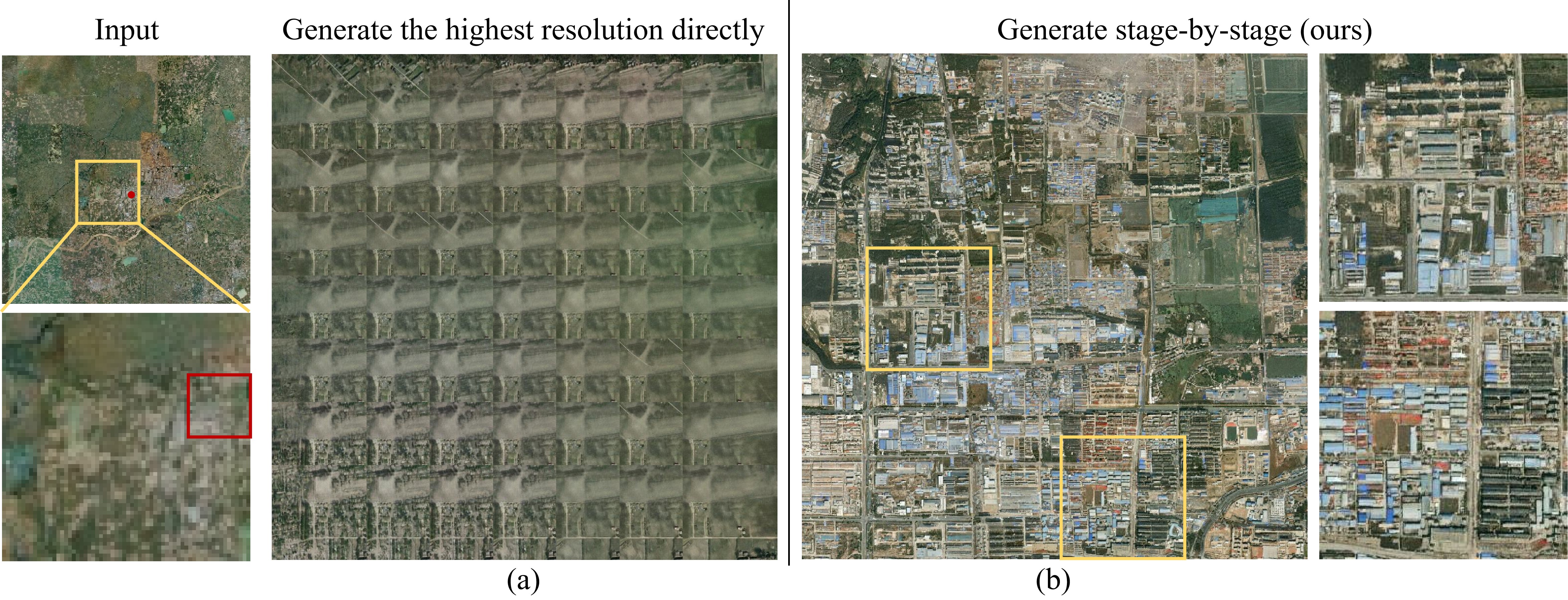}
    \captionsetup{justification=raggedright, singlelinecheck=false}
    \caption{Comparison between directly generating the highest resolution image (a) and generating it in a stage-by-stage manner (b). The generated images have a resolution of 4m/pix and a size of 1024$\times$1024 pixels. With the red-boxed region from the first column of images as input, it is evident that the cascaded generation image exhibits more distinct semantics, clearer features, and a more rational layout than the directly generated one.}
	\label{fig:exp_cascaded_framework}
\end{figure*}

To verify the necessity and rationality of self-cascaded generation framework within the limits of maximum hardware feasibility, we developed a generative model that is structurally and parametrically close to MetaEarth, with an input size of 4$\times$4 and an output size of 256$\times$256, capable of directly generating 4m resolution images from 256m resolution imagery. Employing our simplified unbounded image generation method, the large-sized image with 4m resolution generated directly by this model and the one generated using the MetaEarth cascaded approach are shown in Fig.~\ref{fig:exp_cascaded_framework}.

It is evident that the image generated through a cascaded approach clearly surpasses the directly generated one in visual quality, exhibiting more distinct semantics, sharper feature contours, and a more rational layout. The structural and global semantic information provided by the 4$\times$4-sized images is severely lacking, which will lead to two aspects of problems. On the one hand, this makes it difficult for the model to produce high-quality images with clear features and a sensible distribution of features. On the other hand, this allows the model to be approximated as an unconditional generative model, where the generation results are primarily determined by the initial noise, with the conditional image only having a minor impact. With our simplified unbounded image generation method, the identical initial noise leads to similar outputs across image blocks, resulting in the failure of the algorithm to compile them into a semantically coherent, continuous large-scale image. Considering these points, we therefore conclude that directly generating the highest resolution images is not as sensible an approach, and that employing a cascaded generation framework is essential.

\subsubsection{Effectivene of Resolution Guidance}

In our method, we introduce spatial resolution as a condition to guide the model in perceiving features and details of images at different scales, achieving the generation of images at multiple resolutions using a single model. To evaluate the impact of spatial resolution guidance on the quality of generated images, we design a set of ablation experiments where we solely use the images generated from the previous stage as conditional inputs and compare their image generation quality with the full-implemented model under various scenarios. Specifically, we remove the branches that take resolution as input (frequency encoding $f_\omega$ and MLP network $F_{mlp}$) while keeping the other structures and parameter settings unchanged. During the inference, both models adopt the DDIM acceleration strategy, setting ${\eta=0}$.

We conducted quantitative analysis in different regions around the world. The regions are divided by using different criteria, including the division of regions at different longitudes, or the division of northern and southern hemispheres. The experimental results are shown in Table~\ref{tab:sr-ablation}. The ``w/o sr'' rows represent the network without using the resolution guidance. The first column of the table gives the resolution of the generated images. Except for the 4m*, the model inputs are low-resolution images obtained by downsampling the corresponding resolution real images through bicubic interpolation. The inputs for 4m* are randomly cropped from the real 16m resolution images.  Due to the different characteristics of sensors of different resolution levels, the FID value of 4m* is significantly higher than that of other rows, but our resolution guidance strategy is still effective.

We can see that compared to the model w/o using resolution guidance, our model achieves better FID scores in all testing scenarios except for generating images of the southern hemisphere. In our design, the resolution embedding scales the features of the denoising network, allowing the model to generate images with unique features specific to different spatial resolutions and produce higher-quality outputs. Our experimental results also confirmed this. Such performance improvement becomes more pronounced when the input consists of real images with more complex distributions. The decline in image generation quality for the southern hemisphere is primarily due to the limited variability of land features at different resolutions in this region. The southern hemisphere encompasses vast areas of deserts (in Africa and Oceania) and forests (in South America), where the differences in features across different resolutions are relatively minor compared to the northern hemisphere. Therefore, even without resolution guidance, the model may also generate images of better consistency with the true distribution.

\begin{figure*}
	\centering
	\includegraphics[width=0.95\linewidth]{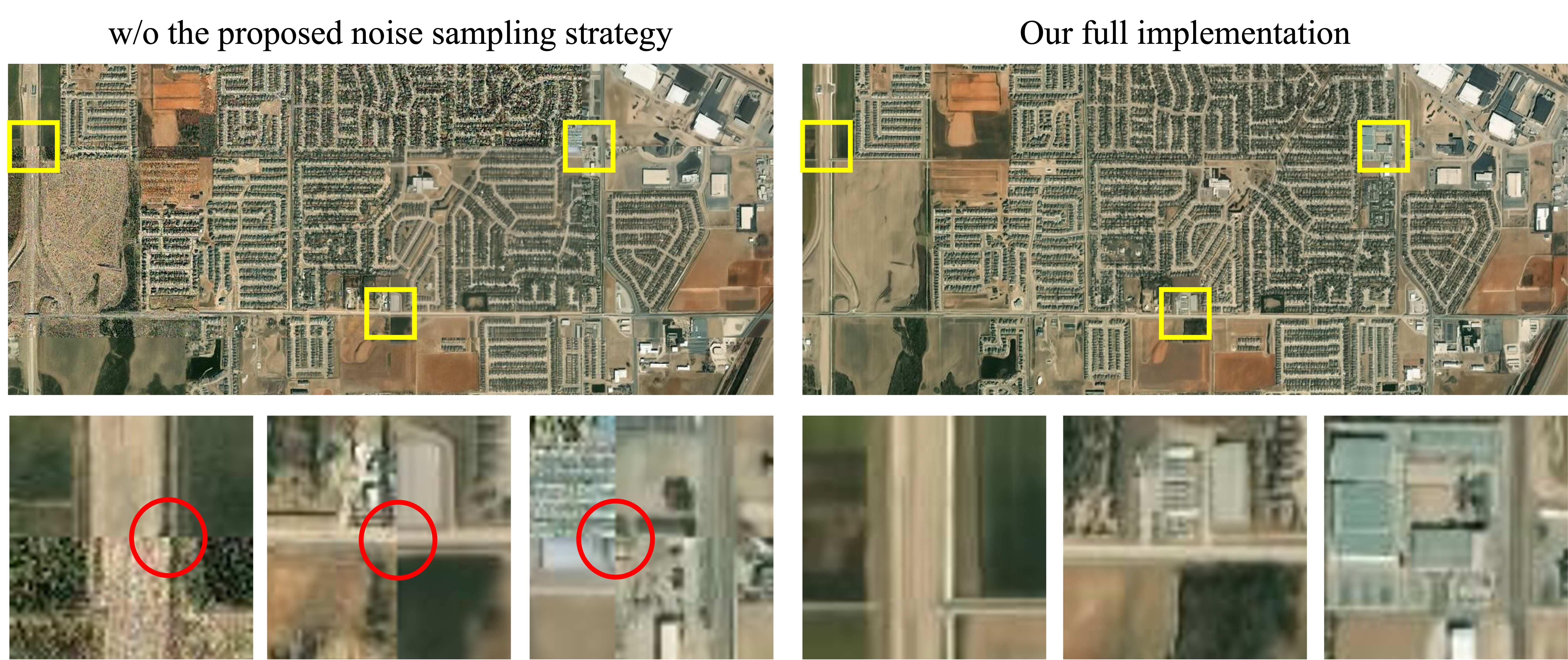}
	\caption{A comparison between the generated images w/ and w/o using the proposed initial noise sampling strategy. It can be seen that without using the noise sampling method proposed in this paper, there are noticeable seams at the junctions of the generated windows, and the semantic content appears discontinuous. In contrast, the proposed method effectively resolves these problems. The images in the second row are local zoomed-in views of the inner regions.}
	\label{fig:exp_seams_compare}
\end{figure*}

\subsubsection{Evaluation on Noise Sampling}

To measure the continuity of the different generated blocks in the seam area, in Fig.~\ref{fig:method_sliding_windows}, we compare the generated results w/ and w/o using the proposed noise sampling strategy. We can see that with the proposed noise sampling strategy, we obtained smoother results, and the traces of patchwork are difficult to detect with the naked eye. We also designed a gradient-based method to quantitatively evaluation the transition smoothness of different approaches. We randomly sample 750 generated images and calculate the absolute directional gradient values of adjacent pixels at the seams and then take the average. Experimental results are shown in Table~\ref{tab:edge-ablation}. We evaluate the proposed two sub-strategies respectively, one is overlapped sliding window (marked as ``Overlap" in Table~\ref{tab:edge-ablation}) and another is the constraints on initial noise sampling (marked as ``Noise Constraint" in Table~\ref{tab:edge-ablation}).  Smaller gradient values indicate better results.  
The experimental results demonstrate that the combined use of the overlapping cropping strategy and our proposed noise sampling method can effectively alleviate the seams appearing during image block stitching, thereby enhancing the visual quality of the generated arbitrary-sized images.

\begin{table}[!ht] 
\renewcommand{\arraystretch}{1.3}
\captionsetup{justification=centering, labelsep=newline, font=small}
\caption{Ablation experiment on our proposed noise sampling method. We compare the averaged directional gradient value of pixels in the window boundary area (a lower value indicates better).}
\label{tab:edge-ablation}
\centering
\begin{tabular}{cc|ccc}
	\toprule
	\multicolumn{2}{c}{Methods} & \multicolumn{3}{c}{${{\rm{Directional Gradient}}\downarrow}$}\\
        {Overlap} & {Noise Constraints} & {Horizontal} & {Vertical} & {Average}\\
    \midrule
        {${\times}$} & {${\times}$} & {24.45} & 24.87 & 24.67 \\ 
        {\checkmark} & {${\times}$} & 17.66 & 18.09 & 17.88 \\
        {\checkmark} & {\checkmark} & ${\boldsymbol{{12.01}}}$ & ${\boldsymbol{{13.11}}}$ & ${\boldsymbol{{12.58}}}$ \\
	\bottomrule
\end{tabular}
\end{table}

\subsubsection{Necessity of 
High-order Degradation Operation}

\begin{figure*}
	\centering
	\includegraphics[width=0.9\linewidth]{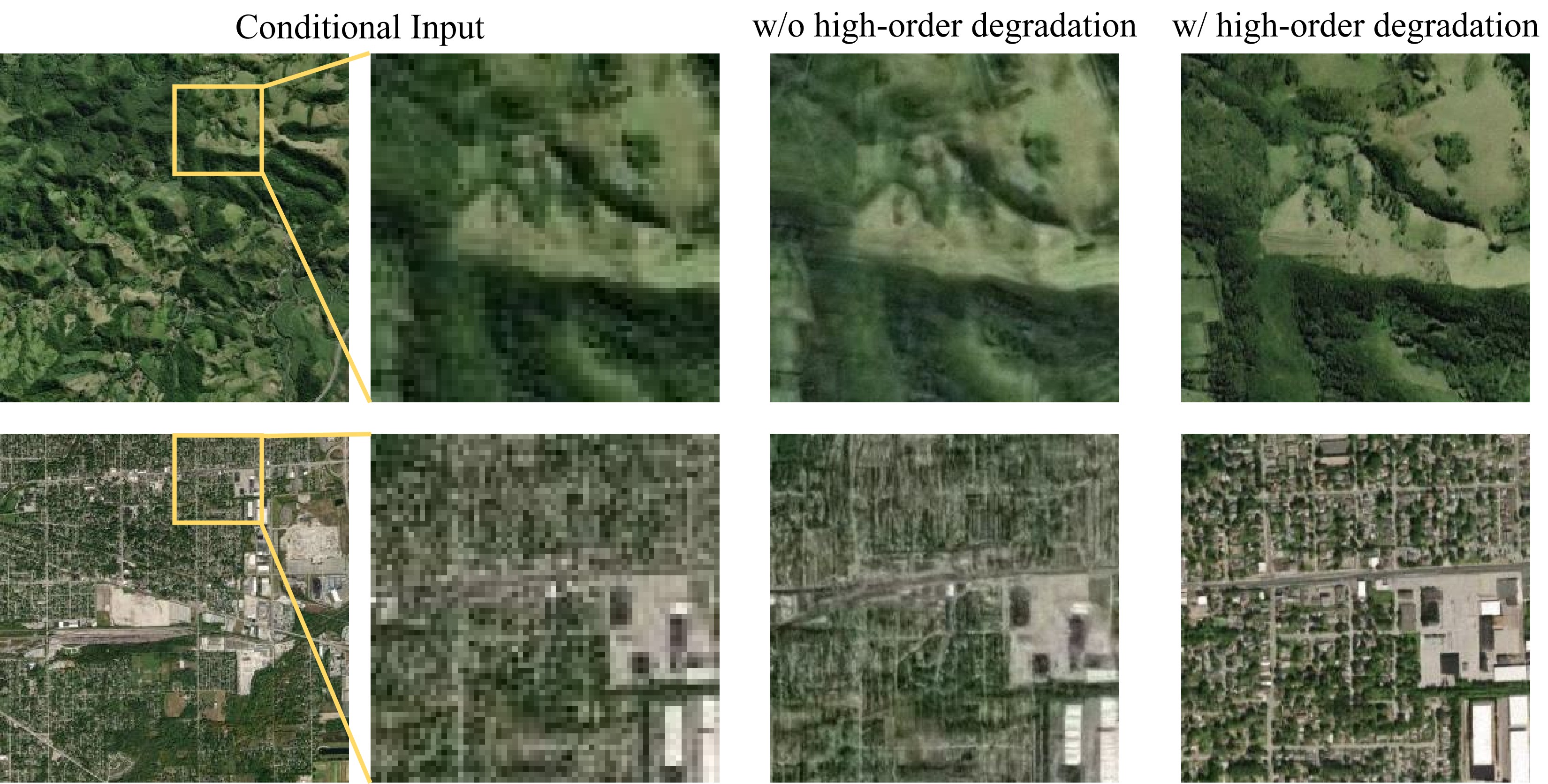}
	\caption{A comparison between the generated images w/ and w/o using the high-order degradation. Both models take real images with a spatial resolution of 16m/pix as input.}
	\label{fig:exp_degradation}
\end{figure*}

In the training process, we introduce high-order degradation operation to address the challenges of shifting between different distributions. In this section, we have designed a new ablation experiment to validate the necessity of the high-order degradation operation, where we compared the simple downsampling method (bicubic interpolation) with the high-order degradation strategy. 

The experiment primarily investigates the single-stage generation process from 16m to 4m resolution, using both bicubic interpolation and high-order degradation to obtain low-resolution images from high-resolution ones for training. Under the same model architecture, parameter volume, training data, and other experimental conditions, we examine the performance of the two generated models in response to different inputs.

As shown in Fig.~\ref{fig:exp_degradation}, when both models process real remote sensing images, the one trained with high-order degradation method generates clearer imagery, whereas the model relying on simple downsampling method results in significantly blurred images. For a well-trained model, the distribution of its output images is similar to that of the real data, which means that the model without high-order degradation cannot take the images generated by the previous stage of the model as input to cascade and generate higher-resolution images, and thus cannot adapt to the self-cascaded framework.

We also calculated the FID for the images generated by the two models. It can be observed from the Table~\ref{tab:degradation-ablation} that our method exhibits a notable advantage in terms of FID, signifying that the high-order degradation we implemented effectively boosts the model's ability to generalize to real input data, resulting in the generation of superior quality high-resolution images.

Both qualitative and quantitative analyses demonstrate that the high-order degradation operation we have adopted is necessary, enhancing the model's adaptability to real data inputs and enabling the model to be applied in a self-cascaded generative framework.

\begin{table}[!ht] 
\renewcommand{\arraystretch}{1.3}
\captionsetup{justification=centering, labelsep=newline, font=small}
\caption{Ablation experiment on high-order degradation operation.}
\label{tab:degradation-ablation}
\centering
\begin{tabular}{m{80pt}<{\centering}m{100pt}<{\centering}}
	\toprule
	{High-order Degradation} & {FID$\downarrow$} \\
	\midrule
        {$\times$} & 148.31\\
        {\checkmark} & ${\boldsymbol{{78.72}}}$\\
	\bottomrule
\end{tabular}
\end{table}

\subsection{Evaluation on Downstream Tasks}

The proposed MetaEarth can be used as a powerful data engine to generate high-quality training data that can benefit downstream tasks. For instance, MetaEarth enables data augmentation by employing a process of downsampling followed by upsampling, providing a vast amount of diverse data to enhance classification tasks that suffer from insufficient or sparse data. In addition, MetaEarth can be utilized to synthesize hybrid scenes that combine real objects and virtual background for object detection, especially for tasks involving rare or scarce targets. Furthermore, MetaEarth may have great potential to build deepfake detection algorithms in the remote sensing field, maintaining the security of geographic information systems and preventing false data from misleading decision-making. With MetaEarth's robust generation capabilities across a global scale and various resolutions, users can obtain multi-scale images with different regional characteristics. This may further enhance the generalization of deepfake detection models and improve their ability to distinguish between real and fake data.

In this paper, to quantitatively demonstrate the application of our model in downstream tasks and further evaluate the fidelity and diversity of the generated images, we choose remote sensing image classification, a fundamental task in remote sensing image interpretation, as our test bed for downstream evaluation. In this experiment, we use our model as a data engine to provide training data support. Our classification dataset contains seven categories of remote sensing scenes: beach, desert, farmland, forest, industrial area, mountain, and residential area. Each class contains approximately 150 images, with image sizes of 256$\times$256 pixels. The dataset is re-collected from our global 4m/pix resolution images. The training and testing sets are divided by a 3:1 ratio.

\begin{figure}
	\centering
	\includegraphics[width=0.45\textwidth]{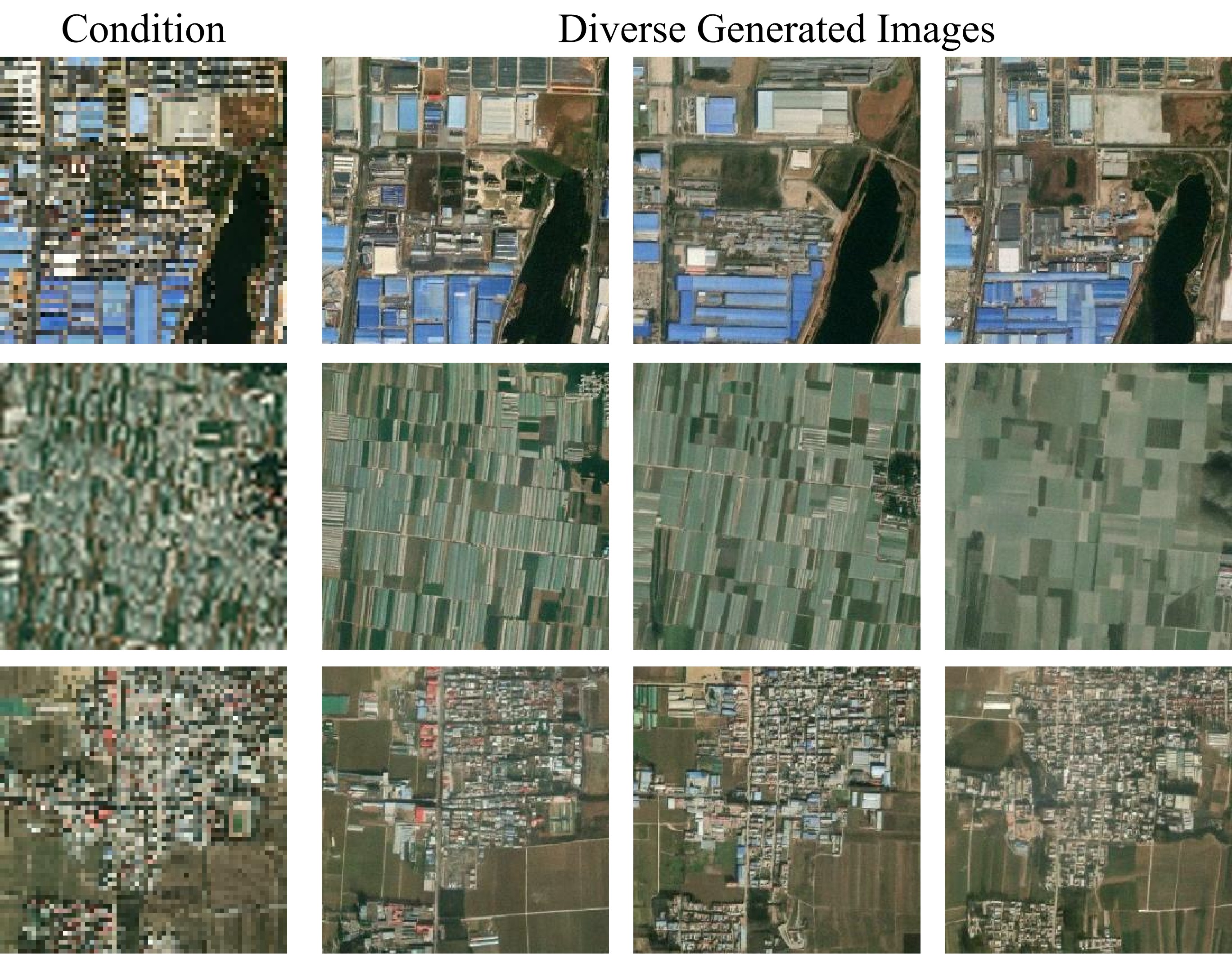}	
	\caption{The generated augmented data. The leftmost column displays input conditional images, with the top-to-bottom rows representing industrial areas, farmlands, and residential areas, respectively. The right three columns show images generated from the real images.}
	\label{fig:exp_classification}
\end{figure} 

We choose four different models as classifiers: VGG19~\cite{vgg}, ResNet34~\cite{resnet}, ViT-B/32, and ViT-B/16~\cite{vit}. We first train four baseline models using the original dataset. Then, we use the generation model to augment the training dataset to five times its original size. After mixing the augmented data with the original one, we retrain the four models from scratch using the combined dataset. For the augmentation process, we downsample the images in the training dataset and utilize the model to generate diverse results. Some of the augmented results are shown in Fig~\ref{fig:exp_classification}.

\begin{table}[!ht] 
\renewcommand{\arraystretch}{1.3}
\captionsetup{justification=centering, labelsep=newline, font=small}
\caption{Accuracy of the downstream remote sensing image classification task w/ and w/o image augmentation.}
\label{tab:classification}
\centering
\begin{tabular}{c|c|c}
	\toprule
	{Methods} & {Traning Data} & {Accuracy}\\
	\midrule
	\multirow{2}{*}{VGG19} & {Real Data} & 95.72\\
    & {Augment Data} & ${\boldsymbol{97.28}}$\\
    \midrule
	\multirow{2}{*}{ResNet34} & {Real Data} & 96.11\\
    & {Augment Data} & ${\boldsymbol{99.22}}$\\
    \midrule
	\multirow{2}{*}{ViT-B/32} & {Real Data} & 93.77\\
    & {Augment Data} & ${\boldsymbol{94.55}}$\\
    \midrule
	\multirow{2}{*}{ViT-B/16} & {Real Data} & 93.77\\
    & {Augment Data} & ${\boldsymbol{95.33}}$\\
	\bottomrule
\end{tabular}
\end{table}

Table~\ref{tab:classification} shows the classification accuracy on the test set for different models using different training data. Compared with the original models, the models trained with data augmentation show consistently improved classification accuracy. This indicates that MetaEarth provides informative data for training, which also reflects the high quality and diversity of the images generated. This indicates that when faced with tasks with insufficient or sparse data, MetaEarth has the potential to serve as a data engine to provide usable data at lower costs.

\subsection{Limitation}

\begin{figure}
	\centering
	\includegraphics[width=0.9\linewidth]{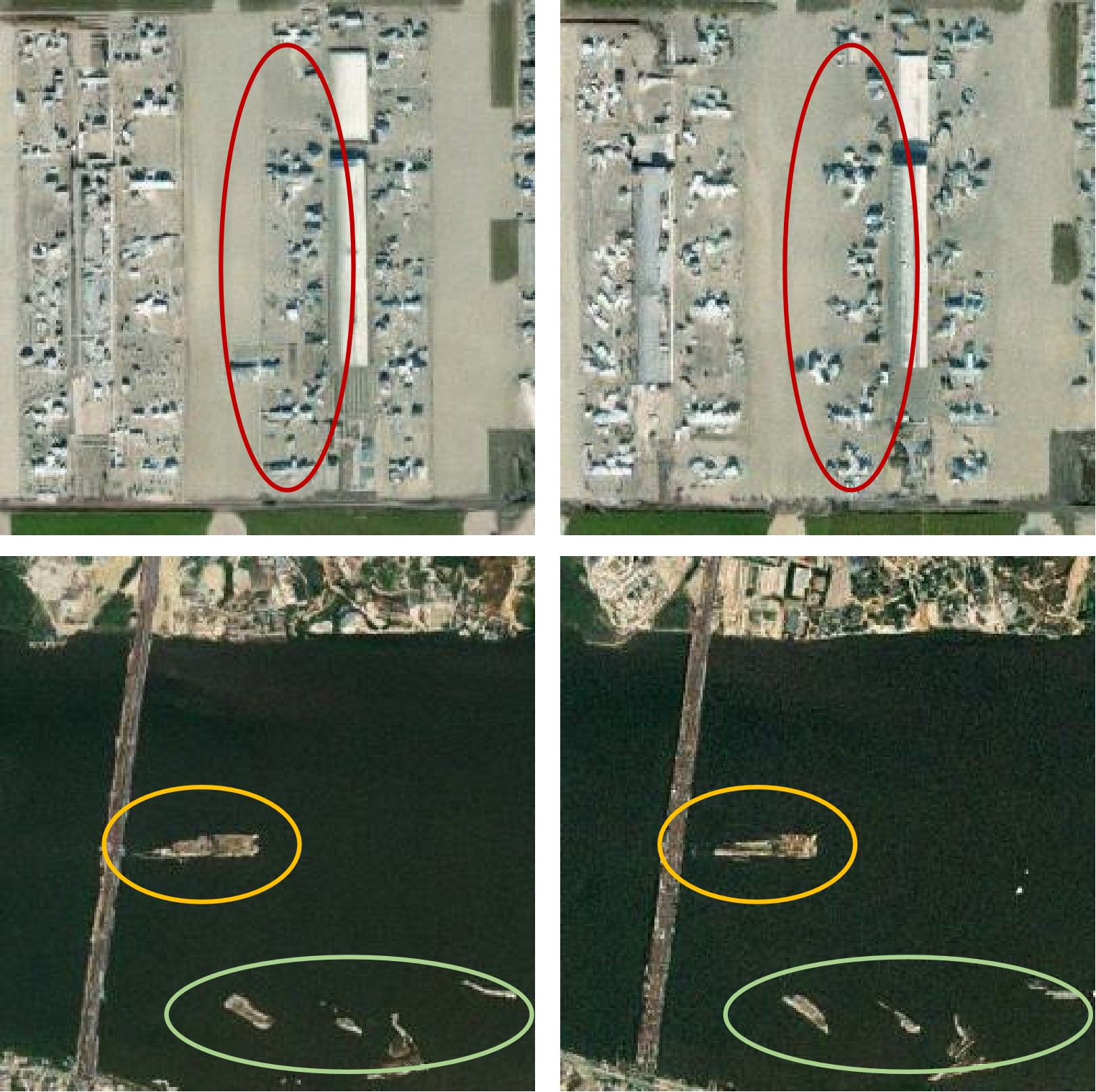}
    \captionsetup{justification=raggedright}
    \caption{Failure cases of MetaEarth in generating small-sized objects. The first and second rows show unsuccessful results in generating airplanes and ships.}
	\label{fig:failure_cases}
\end{figure} 

Currently MetaEarth may not be able to generate small objects such as aircraft and ships well. Some failure cases are shown in Fig.~\ref{fig:failure_cases}, where the shapes of the planes or ships are almost completely lost, and their features are hard to discern. This is due to the fact that, constrained by computational resources, our highest resolution for training data is 4m/pix. At this resolution, the targets are so small and occupy very few pixels, leading to a significant issue of data imbalance. During training, background patterns become the dominant focus of learning, making it challenging for the model to accurately capture the characteristics of the targets. In future research, to address these issues, we will continue to explore the potential of remote sensing generative models to achieve the generation of higher resolution images and controllable embedding of targets. We will also attempt to expand the generated modalities, not limited to the visible light images, and further investigate the controllable generation of multimodal data such as multispectral and DEM images.

\section{Discussion}

In addition to serving as a data engine for data augmentation, MetaEarth also holds immense potential in the construction of generative world models. World models are a crucial pathway to achieving general artificial intelligence, providing essential training and testing environments for an agent's perception, decision-making, and evolution. Recently, the emergence of OpenAI's Sora model has brought significant attention to research in the field of world models. Generative world models seek to understand the world through generative processes, typically integrating closely with technologies such as large language models, visual foundation models, and video/image generation models.

As a generative foundation model, MetaEarth opens up new possibilities for constructing generative world models by simulating Earth’s visuals from an innovative overhead perspective. This unique approach allows for the creation of high-resolution, multi-resolution, and unbounded remote sensing images that cover diverse geographical features and ecological systems. Given a virtual viewpoint’s altitude and movement path, MetaEarth can generate and predict content beyond the immediate scene, thus creating a dynamic environment with intelligent interactive behaviors. By leveraging MetaEarth, researchers may generate a realistic and dynamic visual environment that can interact with intelligent agents such as drones and remote sensing satellites in various scenarios, ranging from urban planning and environmental monitoring to disaster management and agricultural optimization, facilitating their training, testing, and validation. 

Constructing generative world models in the aerospace intelligence domain is an important research trend for the future. In this regard, MetaEarth offers valuable insights and methodologies. We believe that as a starting point for this line of research, MetaEarth will significantly contribute to the development of generative world models for aerospace remote sensing, providing new possibilities for its future advancements.

\section{Conclusion}

In this paper, we introduced MetaEarth, a generative foundation model specifically designed for global-scale remote sensing image generation. MetaEarth extends the capabilities of existing generative models by enabling the production of worldwide, multi-resolution, unbounded, and virtually limitless remote sensing images. The proposed resolution-guided self-cascading generative framework and innovative noise sampling strategy allow MetaEarth to generate high-quality images with diverse geographical features and resolutions, overcoming significant challenges in model capacity, resolution control, and unbounded image generation. Extensive experiments demonstrate MetaEarth's powerful capabilities in generating diverse and high-fidelity remote sensing images across different resolutions and regions. It also shows great potential as a data engine, capable of generating high-quality training data for downstream tasks in remote sensing. The advancements presented in this paper pave the way for future research in building generative world models from an overhead perspective of the earth.


%





\ifCLASSOPTIONcaptionsoff
  \newpage
\fi



\bibliographystyle{IEEEtran}
\bibliography{references.bib}

\begin{IEEEbiography}[{\includegraphics[width=1in,height=1.25in,clip,keepaspectratio]{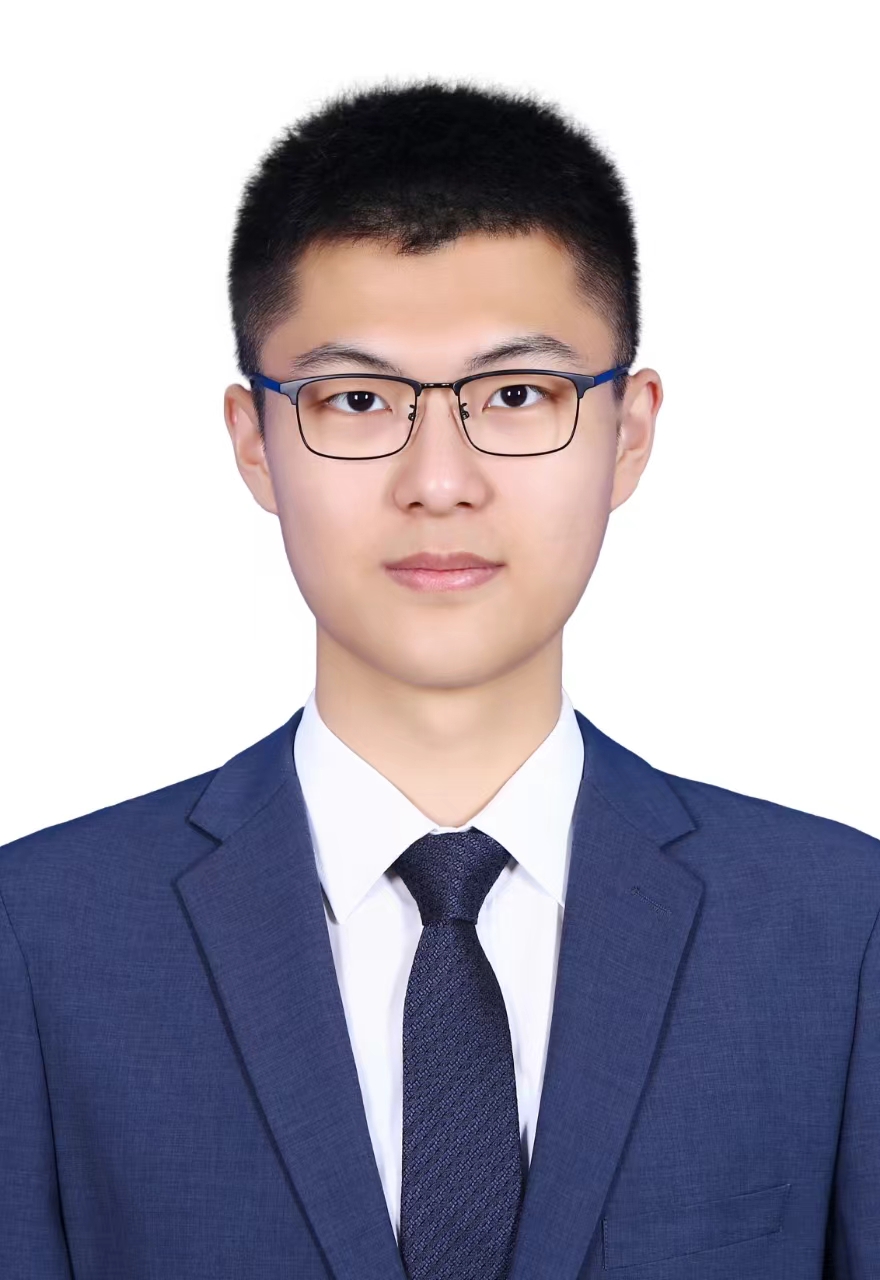}}]
{Zhiping Yu}
is currently an undergraduate student at the Image Processing Center, School of Astronautics, Beihang University. His research interests include deep learning and remote sensing image processing.
\end{IEEEbiography}

\begin{IEEEbiography}[{\includegraphics[width=1in,height=1.25in,clip,keepaspectratio]{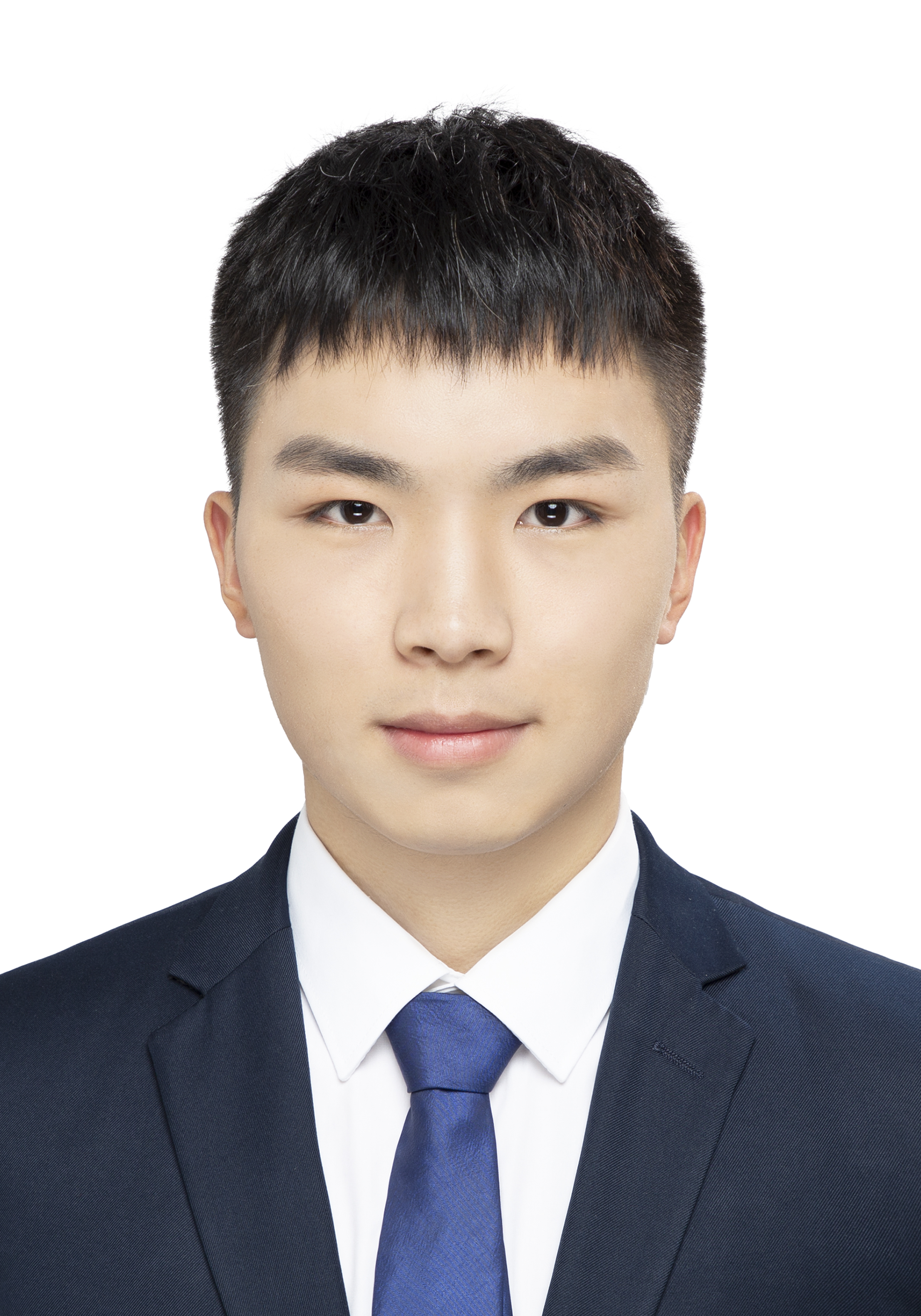}}]
{Chenyang Liu}
received his B.S. degree from the Image Processing Center, School of Astronautics, Beihang University in 2021. He is currently working towards the Ph.D. degree in the Image Processing Center, School of Astronautics, Beihang University. His research interests include machine learning, computer vision, and multimodal learning.
\end{IEEEbiography}

\begin{IEEEbiography}[{\includegraphics[width=1in,height=1.25in,clip,keepaspectratio]{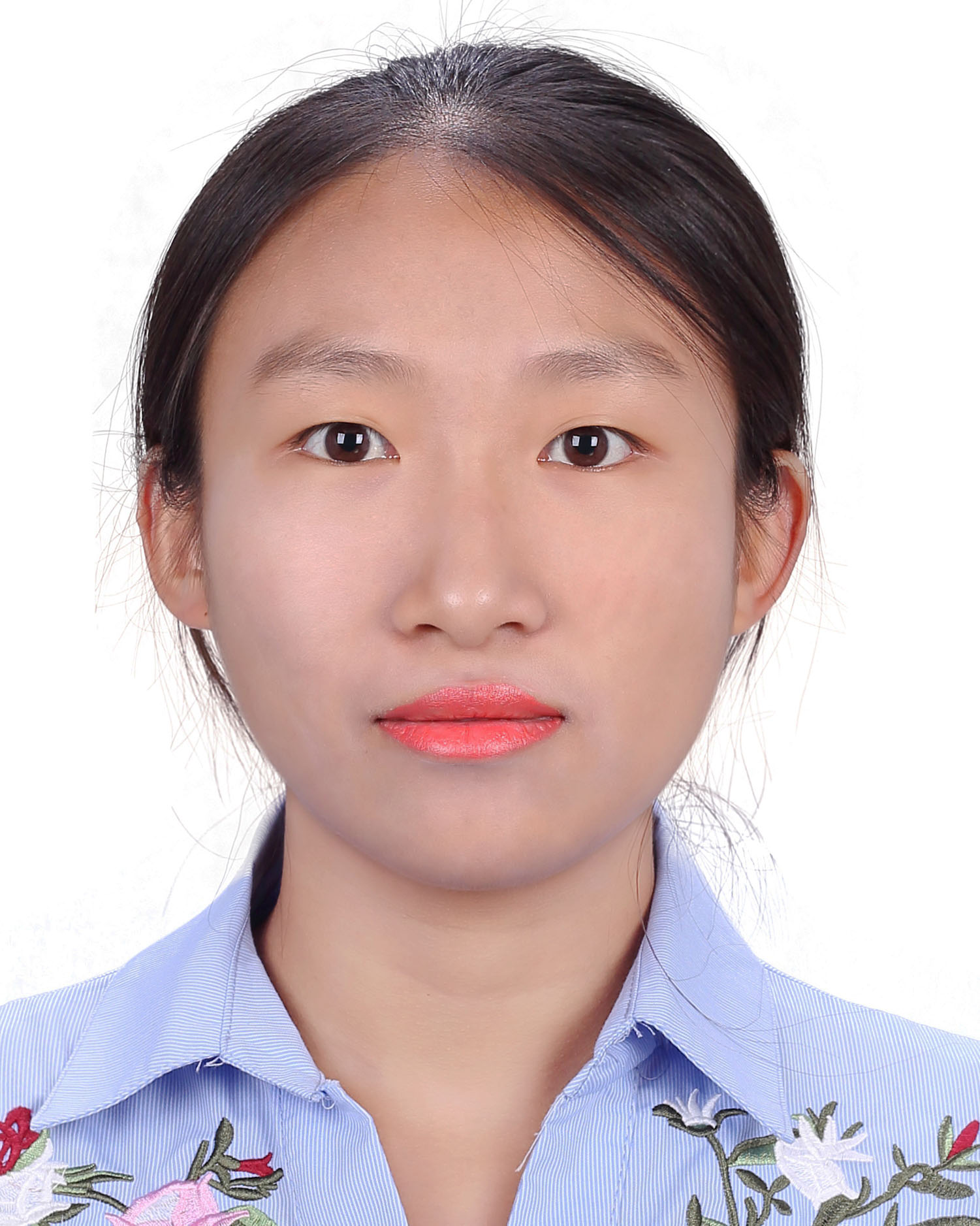}}]
{Liqin Liu}
received her B.S. degree from Beihang University, Beijing, China in 2018. She is currently working toward her doctorate degree in the Image Processing Center, School of Astronautics, Beihang University. Her research interests include hyperspectral image processing, machine learning, and deep learning.
\end{IEEEbiography}

\begin{IEEEbiography}
[{\includegraphics[width=1in,height=1.25in,clip,keepaspectratio]{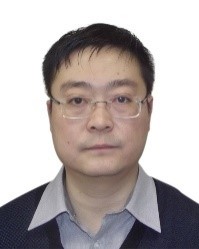}}]
{Zhenwei Shi}
(Senior Member, IEEE) is currently a Professor and Dean of the Image Processing Center, School of Astronautics, Beihang University. He has authored or co-authored over 200 scientific articles in refereed journals and proceedings, including the IEEE Transactions on Pattern Analysis and Machine Intelligence, the IEEE Transactions on Image Processing, the IEEE Transactions on Geoscience and Remote Sensing, the IEEE Conference on Computer Vision and Pattern Recognition (CVPR) and the IEEE International Conference on Computer Vision (ICCV). His current research interests include remote sensing image processing and analysis, computer vision, pattern recognition, and machine learning.

Prof. Shi serves as an Editor for IEEE Transactions on Geoscience and Remote Sensing, Pattern Recognition, ISPRS Journal of Photogrammetry and Remote Sensing, Infrared Physics and Technology, etc. His personal website is http://levir.buaa.edu.cn/.
\end{IEEEbiography}

\begin{IEEEbiography}[{\includegraphics[width=1in,height=1.25in,clip,keepaspectratio]{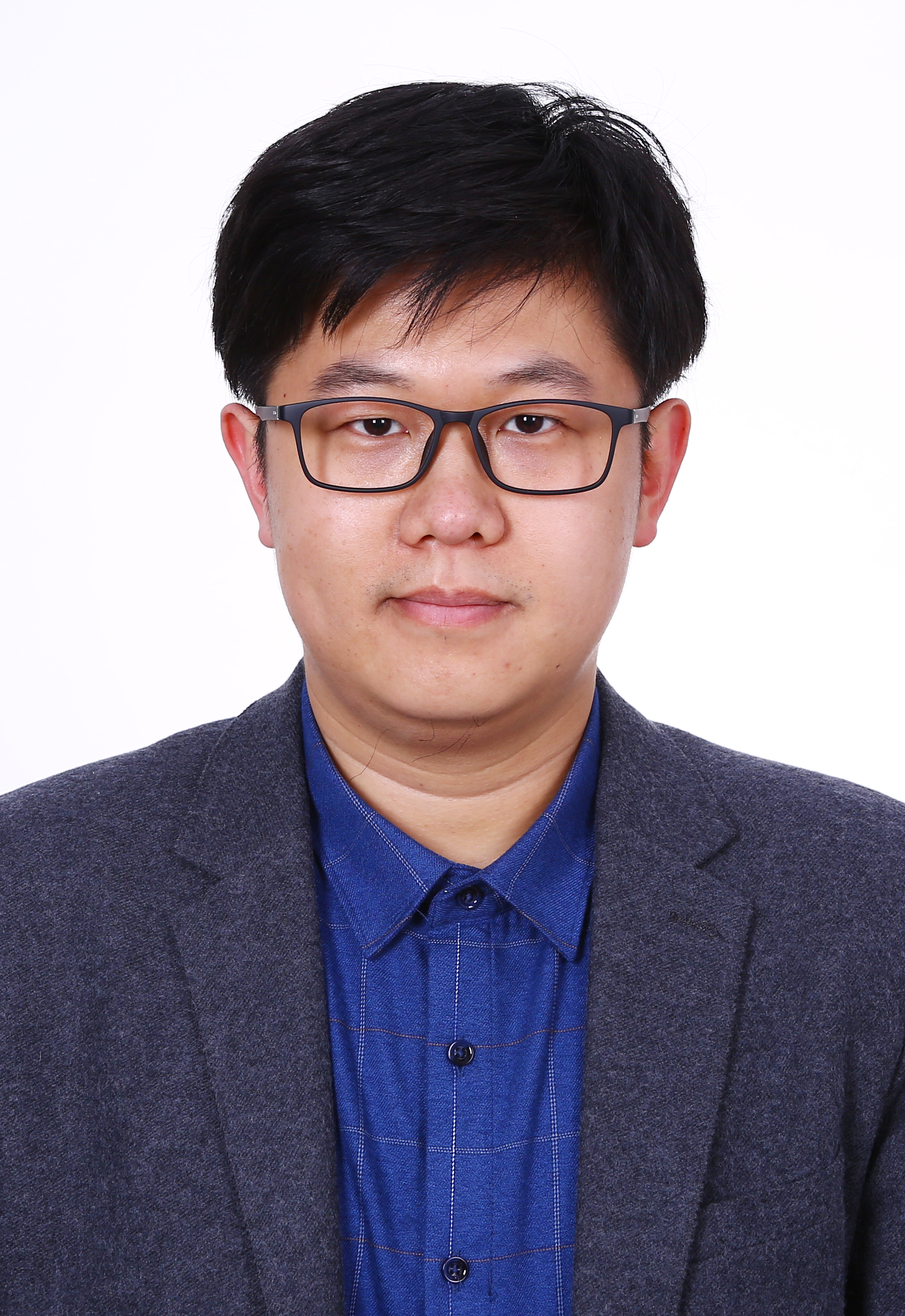}}]{Zhengxia Zou}
(Member, IEEE) received his BS degree and his Ph.D. degree from Beihang University in 2013 and 2018. He is currently a Professor at the School of Astronautics, Beihang University. During 2018-2021, he was a postdoc research fellow at the University of Michigan, Ann Arbor. His research interests include computer vision and related problems in remote sensing. He has published over 30 peer-reviewed papers in top-tier journals and conferences, including Proceedings of the IEEE, Nature Communications, IEEE Transactions on Pattern Analysis and Machine Intelligence, IEEE Transactions on Geoscience and Remote Sensing, and IEEE / CVF Computer Vision and Pattern Recognition. Dr. Zou serves as the Associate Editor for IEEE Transactions on Image Processing. His personal website is \url{https://zhengxiazou.github.io/}.
\end{IEEEbiography}

%








\end{document}